\documentclass[10pt,journal,compsoc]{IEEEtran}

\ifCLASSOPTIONcompsoc
  \usepackage[nocompress]{cite}
\else
  \usepackage{cite}
\fi
\usepackage{ragged2e} 
\usepackage[table]{xcolor}
\usepackage{graphicx}
\usepackage{amssymb}
\usepackage{amsmath}
\usepackage{multirow}
\usepackage{tabularx}
\usepackage{dsfont}
\usepackage{url}
\usepackage{color}
\usepackage{amsthm}
\usepackage[ruled,linesnumbered]{algorithm2e}
\usepackage[export]{adjustbox}
\usepackage{booktabs}
\usepackage{subcaption}
\usepackage[font=footnotesize,labelfont=bf]{caption}
\usepackage{longtable}
\usepackage{tabu}
\usepackage{diagbox,booktabs}
\usepackage{multicol}

\usepackage[numbers]{natbib}

\usepackage{pifont}

\definecolor{darkgreen}{rgb}{0,0.694,0.298}
\definecolor{purple}{rgb}{0.4,0.176,0.569}
\definecolor{royalblue}{RGB}{65,105,225}
\definecolor{americanrose}{rgb}{1.0, 0.01, 0.24}
\definecolor{applegreen}{rgb}{0.55, 0.71, 0.0}

\newcommand{\figref}[1]{Fig.~\ref{#1}}
\newcommand{\reqref}[1]{Eq.~\eqref{#1}}
\newcommand{\secref}[1]{Sec.~\ref{#1}}
\newcommand{\tableref}[1]{Table~\ref{#1}}

\makeatletter
\DeclareRobustCommand\onedot{\futurelet\@let@token\@onedot}
\def\@onedot{\ifx\@let@token.\else.\null\fi\xspace}
\def\eg{\emph{e.g}\onedot} 
\def\ie{\emph{i.e}\onedot} 
 
\def\etc{\emph{etc}\onedot} 
\def\wrt{w.r.t\onedot} 
\def\etal{\emph{et al}\onedot}
\makeatother

\definecolor{americanrose}{rgb}{1.0, 0.01, 0.24}
\definecolor{gray}{rgb}{0.1, 0.1, 0.1}

\newcommand{\first}[1]{\textbf{\textcolor{black}{#1}}}
\newcommand{\second}[1]{\textcolor{gray}{#1}}

\newcommand{\ourmethod}{\textit{EfDeRain+}}

\definecolor{royalblue}{RGB}{65,105,225} 
\usepackage[pagebackref=false,breaklinks=true,colorlinks,citecolor=royalblue,bookmarks=false]{hyperref}

\begin{document}
%

\title{Uncertainty-Aware Cascaded Dilation Filtering for High-Efficiency Deraining}

%
%
%

\author{Qing~Guo,~\IEEEmembership{Member,~IEEE,}
        Jingyang~Sun,
        Felix~Juefei-Xu,~\IEEEmembership{Member,~IEEE,}
        Lei~Ma,~\IEEEmembership{Member,~IEEE,} 
        Di~Lin,~\IEEEmembership{Member,~IEEE,}
        Wei~Feng,~\IEEEmembership{Member,~IEEE,}
        Song Wang,~\IEEEmembership{Senior Member,~IEEE}
\IEEEcompsocitemizethanks{\IEEEcompsocthanksitem Qing Guo is with Nanyang Technological University, Singapore. Jingyang Sun is with Kyushu University, Japan. Lei Ma is with University of Alberta, Canada. Felix Juefei-Xu is with Alibaba Group, USA. Di Lin and Wei Feng are with the School of Computer Science and Technology, College of Intelligence and Computing, Tianjin University, Tianjin 300305, China. Song Wang is with Department of Computer Science and Engineering, University of South Carolina, Columbia, SC 29208, USA.
\protect\\
}
\thanks{Manuscript received April 19, 2005; revised August 26, 2015.}}

%
%

\markboth{Journal of \LaTeX\ Class Files,~Vol.~14, No.~8, August~2015}%
{Shell \MakeLowercase{\textit{et al.}}: Bare Demo of IEEEtran.cls for Computer Society Journals}
%


\IEEEtitleabstractindextext{%
\begin{abstract}
\justifying
Deraining is a significant and fundamental computer vision task, aiming to remove the rain streaks and accumulations in an image or video captured under a rainy day.
Existing deraining methods usually make heuristic assumptions of the rain model, which compels them to employ complex optimization or iterative refinement for high recovery quality.  
This, however, leads to time-consuming methods and affects the effectiveness for addressing rain patterns deviated from from the assumptions. 
In this paper, we propose a simple yet efficient deraining method by formulating deraining as a predictive filtering problem without complex rain model assumptions. 
Specifically, we identify \textit{spatially-variant predictive filtering (SPFilt)} that adaptively predicts proper kernels via a deep network to filter different individual pixels.
Since the filtering can be implemented via well-accelerated convolution, our method can be significantly efficient. 
We further propose the \textit{EfDeRain+} that contains three main contributions to address residual rain traces, multi-scale, and diverse rain patterns without harming the efficiency. \textit{First}, we propose the \textit{uncertainty-aware cascaded predictive filtering (UC-PFilt)} that can identify the difficulties of reconstructing clean pixels via predicted kernels and remove the residual rain traces effectively.
\textit{Second}, we design the \textit{weight-sharing multi-scale dilated filtering (WS-MS-DFilt)} to handle multi-scale rain streaks without harming the efficiency. 
\textit{Third}, to eliminate the gap across diverse rain patterns, we propose a novel data augmentation method (\ie, \textit{RainMix}) to train our deep models.
By combining all contributions with sophisticated analysis on different variants, our final method outperforms baseline methods on four single-image deraining datasets and one video deraining dataset in terms of both recovery quality and speed. In particular, \ourmethod{} can derain within about 6.3~ms on a $481\times 321$ image and is over 74 times faster than the top baseline method with even better recovery quality.
\end{abstract}

}

\maketitle
\IEEEdisplaynontitleabstractindextext

%
\IEEEpeerreviewmaketitle


\IEEEraisesectionheading{\section{Introduction}\label{sec:introduction}}

\IEEEPARstart{R}{ain} patterns (\eg, rain streaks) captured by outdoor computer vision systems (\eg, stationary image or dynamic video sequence), often lead to sharp intensity fluctuations in images or videos, causing performance degradation for the visual perception systems \cite{garg2005does,garg2007vision} across different tasks, \eg, pedestrian detection \cite{Wang2018cvpr},  object tracking \cite{Li18}, semantic segmentation \cite{MaCoSNet}, \etc.
As a common real-world phenomenon, it is almost mandatory that an all-weather vision system is equipped with the deraining capability for usage. A deraining method processes the rain-corrupted image/video data and removes the rain patterns, with the intention to achieve good image quality for the downstream vision tasks or human perception. 

Recent works have achieved significant progresses on public deraining datasets \cite{yang2017cvpr, fu2017removing, wang2019spatial} by studying physical models of rain and background layers to utilize rain pattern priors like similar local shapes, thickness, and directions in a rainy image \cite{kang2012automatic,sun2014exploiting,luo2015removing,li2016rain,gu2017joint,yang2019joint,wang2020a,yang2020single}. As a result, these methods can build the physical model-based objective functions and formulate the deraining as an optimization problem that is solved by iterative optimization or refinement. For example, the state-of-the-art (SOTA) rain convolutional dictionary network (RCDNet) \cite{wang2020a} represents the rainy image via a concise rain convolutional dictionary model and design an iterative optimization method to solve the rain maps and background layer (\ie, the rain-free image).
RCDNet achieves top recovery quality on several benchmarks. 

Although the model-driven deraining methods achieve impressive results, some caveats of theirs cannot be overlooked: \textit{First}, the physical model assumptions adopted by many of the algorithms may be limited and fall short to cover and reflect the real-world rain patterns. The models based on these rain model assumptions may not perform as strongly under the real-world scenarios, which leaves some room for further improvements on the recovery quality; \textit{Second}, many of the existing methods are computationally expensive, requiring complex iterative optimization to find the optimal solution. For example, the RCDNet takes on average 468~ms per image for Rain100H dataset \cite{yang2017cvpr}, which becomes less effective for efficiency-sensitive applications. 
In many real-time applications (\eg, vision-based autonomous driving or navigation), being able to perform deraining \emph{efficiently} on-chip is of great importance. A deraining algorithm achieving both high efficiency and high recovery quality (\eg, in terms of PSNR, SSIM), while remaining at low overhead, is of great importance for practical usage.

With the rapid development of deep learning in low-level vision tasks, a series of data-driven deep deraining methods are proposed and is usually more efficient than the model-driven methods \cite{fu2017removing,li2018recurrent,ren2019cvpr,wang2019spatial,liu2021unpaired}. 
These works design novel deep architecture for deraining and train the network parameters via rainy/clean image pairs. 
Nevertheless, existing data-driven methods also encounter some limitations: \textit{First}, their recovery quality is still worse than the model-driven methods (\eg, RCDNet). One of the critical reasons is the small scale of the training dataset that may not cover complex real rain patterns. \textit{Second}, to reach competitive performance, most of the data-driven methods usually employ recurrent or progressive architectures \cite{ren2019cvpr,li2018recurrent}, neglecting the effects to efficiency.

In this work, we propose a simple yet efficient deraining method by formulating deraining as a predictive filtering problem and make several contributions to improve this intuitive idea. We denote our final method as \textit{EfficientDerain Plus} with \ourmethod{} as a reformation of our conference work \cite{guo2021efficientderain}. 
Specifically, we conduct filtering on each pixel of a rainy image via an exclusive kernel and the clean pixel is reconstructed through linearly weighted sum of the neighboring pixels. The kernel determines the combination weights. 
To allow the weights to be spatially adaptive, we identify the \textit{spatially-variant predictive filtering (SPFilt)} that adaptively predicts proper kernels via a deep network for different pixels. 
Since the filtering can be implemented through convolution that are computationally-accelerated by all deep learning platforms, our method can be significantly efficient.
To further enhance the capability of the intuitive idea to eliminate residual rain traces, address multi-scale rain streaks, and diverse rain patterns, we make three main contributions:
We first propose the \textit{uncertainty-aware cascaded predictive filtering (UC-PFilt)} that can identify the difficulties of recovering rain-free pixels via predicted kernels and refine the deraining results effectively.
In addition, we propose the \textit{weight-sharing multi-scale dilated filtering (WS-MS-DFilt)} to simultaneously address light and heavy rain streaks automatically without harming the efficiency. 
To address the small-scale issue of the training dataset, we propose a novel data augmentation for deraining task, which is denoted as \textit{RainMix}.
Different from the general augmentation technique for image classification, \textit{RainMix} is to produce diverse rain patterns and backgrounds, respectively, instead of augmenting the rainy images directly as a whole. 
As a result, our deraining deep network can see diverse rain scenes and achieve high recovery quality on the testing datasets.
To validate the effectiveness of all the contributions, we conduct extensive ablative studies on different variants of each component, helping us design our final implementation.   
We compare \ourmethod{} with eighteen baseline methods on four single-image deraining datasets and a video deraining dataset.
Our method outperforms most of SOTAs (\eg, PReNet \cite{ren2019cvpr}, SPANet \cite{wang2019spatial}, and RCDNet \cite{wang2020a}) with an impressive high speed (\eg, 6.3~ms on a $481\times381$ image).

\begin{figure}[t]
	\centering
	\includegraphics[width=1.0\columnwidth]{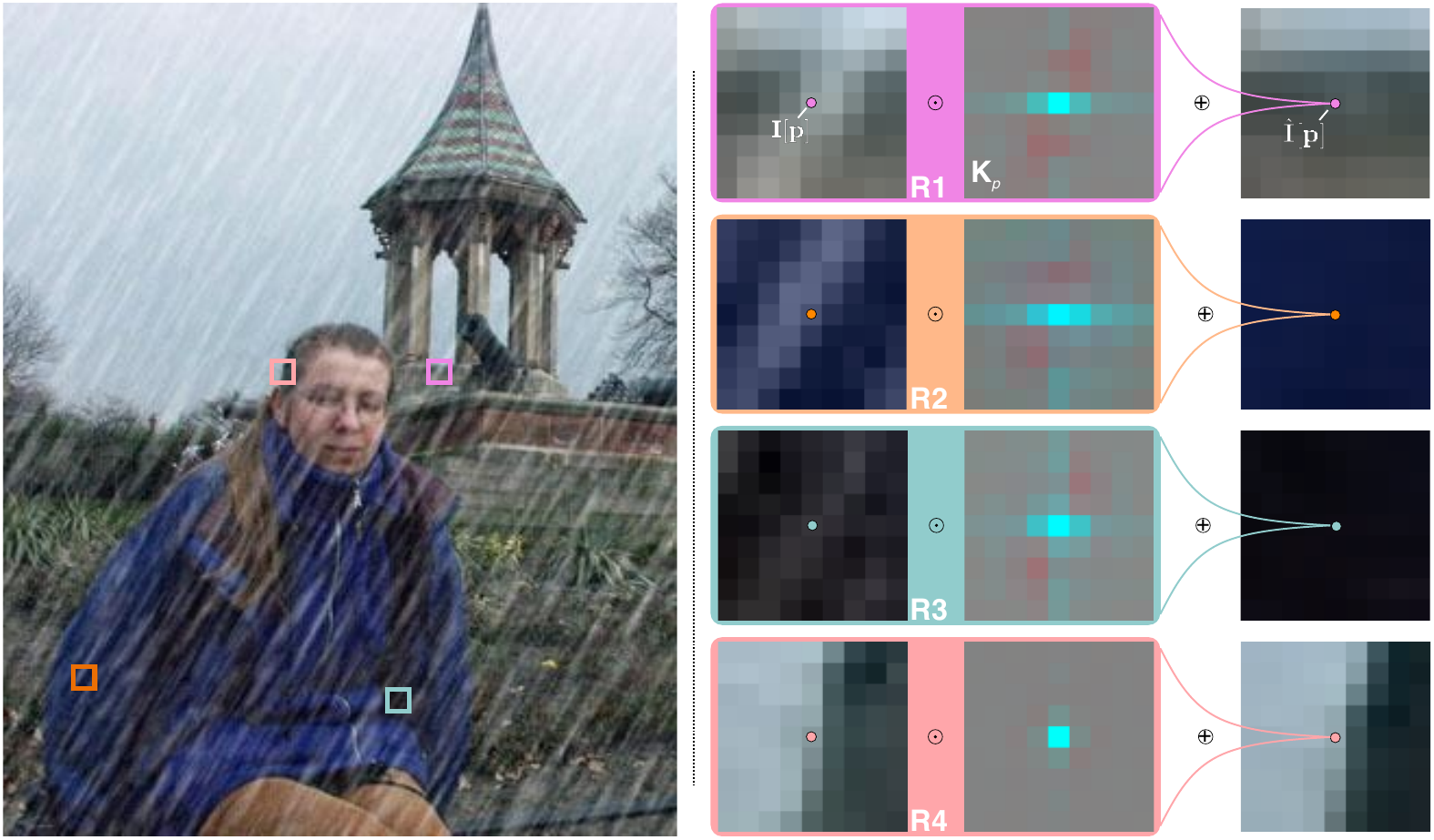}
	\vspace{-20pt}
	\caption{Examples of using spatially-variant filtering to remove rain streaks. Four $9\times 9$ patches  (\ie, R1, R2, R3, and R4) are selected from the rainy image on the left. We process the center pixels within each patch via the predicted kernels: performing pixel-wise multiplication between the patch and a $9\times9$ kernel and adding them together. We use the summation result to replace the original center pixel. By handling all pixels in the same way, we can remove the rain. The key problem is how to generate proper kernels for different pixels.}
	\vspace{-15pt}
	\label{fig:imgfilter}
\end{figure}

\section{Related Work}

\subsection{Single-image Deraining}

Early methods adopt the rain model-based solutions and regard the single-image deraining as an image decomposition task \cite{kang2012automatic,sun2014exploiting,luo2015removing,li2016rain,gu2017joint,wang2017hierarchical,zhu2017joint,wang2020a}. The main goal is to decompose a rainy image to different parts (\eg, low-frequency and high-frequency signals, or large-scale structures and fine-scale structures) and recover the background image by solving an optimization problem with some intuitive priors as regularization terms. For example, Kang \etal{} \cite{kang2012automatic} assume rain information is included in the high-frequency part of a rainy image. They decompose a rainy image into high-frequency and low-frequency parts and model the decomposition task based on the morphological component analysis with a novel objective function. The background image is restored via iterative optimization. Subsequent works adopt similar ideas but with different rain models or optimization methods. Specifically, Sun \etal{} \cite{sun2014exploiting} utilize the structure similarity across rain streaks for better clean-pixel reconstruction. Luo \etal{} \cite{luo2015removing} proposes a non-linear composite rain model and conduct deraining via a greedy pursuit algorithm. Li \etal{} \cite{li2016rain} assume the additive rain layer and use two Gaussian mixture models to represent priors of the rain layer and background image, respectively. Gu \etal{} \cite{gu2017joint} propose the joint analysis sparse representation and synthesis sparse representation to model the image decomposition problem and employ the alternating direction method of multipliers (ADMM) for iterative optimization. 
In general, the image decomposition-based deraining has achieved significant progresses and contributed the community with a series of original methods. Nevertheless, they can hardly cover diverse rain patterns in the real world and some residual rains cannot be properly addressed.

To achieve high-quality restoration, a series of works borrow the power of deep learning for effective deraining with novel deep architectures or modules \cite{fu2017tip,fu2017removing,li2018recurrent,zhang2018iccv,yang2019joint,wang2019spatial,cvid,wang2020a,chen2021robust,zhou2021image,xiao2021improving,huang2021memory,wang2021multi,liu2021unpaired,yi2021structure}. Specifically, Yang \etal{} \cite{yang2019joint} propose a new rain model by considering the global atmospheric light to cover the rain streak accumulation and direction variations. Then, they design a contextualized dilated network to identify rain regions and predict the background recurrently. %
Li \etal{} \cite{li2018recurrent} further generalize the rain model by assigning different atmospheric parameters to multiple rain layers, and propose a recurrent structure to iteratively decompose rain streaks.
More recently, Wang \etal{} \cite{wang2020a} model the rain layer via a convolutional dictionary with pre-defined rain streak priors and propose a model-driven deep neural network for deraining iteratively. 
Overall, many state-of-the-art deep learning-based deraining methods build more complex rain models to cover diverse rain patterns and employ recurrent or iterative strategies to overcome the limitations of traditional image decomposition-based deraining solution. Nevertheless, it is hard to say the new rain models could represent all rain patterns that may happen in the real world, and the iterative strategy introduces limited improvement on the efficiency.

We argue that existing deep learning-based deraining methods have not unleashed the full potential of deep learning such as highly efficient inference as well as strong generalization capability. This is partially due to the aforementioned complex iterative optimizations required, dependencies on strong prior knowledge such as various rain models, as well as scarcity of real-world rain data. Our proposed method fixes these issues by formulating deraining as a predictive filtering task with highly efficient inference. Moreover, with a novel designed data augmentation technique, the trained network presents high generalization across diverse rain patterns.

\subsection{Video and Multi-image Deraining}

There are also a string of work that perform deraining based on video by incorporating information across multiple frames \cite{kim2015video,ren2017video,yang2019frame,yang2020self}.
The main idea for video deraining is to identify the rain regions via temporal information (\eg, optical flow or temporal consistency) and then leverage the identifications to separate the background frame from the rainy frame through advanced optimization methods (\eg, low rank) or pre-trained deep neural networks.
%
%
Nevertheless, under the scenarios where there are drastic changes in the scene (\eg, camera shake), the cross-frame continuity assumption might no longer hold, leading towards inferior deraining performance. Interestingly, we have observed that it is actually feasible to carry out video-based deraining with on-par or even better restoration quality using our single-frame deraining method as we will show in the experimental section.

\subsection{Filtering-based Image Restoration}

Filtering techniques are very widely used in image processing. Due to the local coherence, continuity, and structures existed in natural images, it make sense to process the image locally using filtering techniques. Filtering is considered one of the most prominent techniques for digital image restoration, among many other traditional techniques ranging from classical signal processing methods to estimation theory, and numerical analysis \cite{banham1997digital}. Over the years, many advanced filtering techniques have been developed to tackle various image processing tasks such as adaptive image filtering \cite{awate2006unsupervised}, recursive filtering \cite{kundur1998novel}, multi-resolution bilateral filtering for image denoising \cite{zhang2008multiresolution}, non-local filtering for image denoising \cite{buades2005non}, global filtering for image denoising \cite{talebi2013global}, tri-state median filter for image denoising \cite{chen1999tri}, sparse 3-D transform-domain collaborative filtering for image denoising \cite{dabov2007image}, \etc. 
%
In particular, a non-local means filtering method \cite{kim2013singleimage} is proposed for deraining by first detecting the rain streak regions based on analysis of rotation angle and aspect ratio of the elliptical kernels at each pixel, and then by selecting non-local neighboring pixels adaptively to perform filtering. 
However, this work only utilizes the local context information around rainy pixels and is limited to address some typical rain streaks.

More recently, with the advent of deep neural networks, learning-based predictive filtering starts to trend as a very effective image processing tool by capitalizing the large-scale paired training data in a supervised fashion, as well as the high representation power of a deep network. Among which, the kernel prediction network (KPN) \cite{BakoACMTOG2017,MildenhallCVPR2018} is one such seminal work. KPN is originally proposed to tackle the burst image denoising problem. The gist is to learn per-pixel predictive filters through a deep convolutional neural network, and the learned filters change accordingly with different input images during the inference. These image-aware filters process the input images to produce the final output, \ie, a denoised version. Due to the flexibility of being able to adapt to different input images, and the per-pixel filtering capability, KPN has been widely used beyond image denoising for other image, video, and multi-media restoration tasks \cite{lee2021iterative,yue2020supervised,zhang2021dynamic,pan2020cascaded,fu2021auto}. 
%
%
In a nutshell, many of the exiting works that incorporate predictive filtering, especially for the image restoration tasks \cite{fu2021auto,bhat2021deep}, do not quite consider multi-scale modeling of the problem, and when they do \cite{jiang2021multi}, extra model parameters are usually incurred. In addition, previous works do not assess the capability of the predicted kernels, which could lead to distorted details. 
As a comparison, we equip the predictive filtering with three new modules. First, we propose a novel multi-scale module that improves the deraining quality without harming efficiency. Second, we use a self-produced an uncertainty map to guide the filtering and refine the residual rain trace. Finally, we propose a novel data augmentation for training a more powerful predictive filtering.

\subsection{Data Augmentation Techniques}

Data augmentation were primarily based on geometric and photometric changes of the images in the early years of the deep learning era. More recently, more advanced data augmentation are emerging that brings various advantages to modern era deep learning such as improved generalization capability, reduced domain gap, \etc \cite{yun2019cutmix,devries2017improved,walawalkar2020attentive,zhang2017mixup,chen2020gridmask,lee2020smoothmix,hendrycks2020augmix,xu2020robust,wang2021augmax}. 
In particular, to improve the robustness and uncertainty estimates of image classifiers, AugMix \cite{hendrycks2020augmix} samples various augmentation paths, each with various number of preset augmentations, and the final augmentation is produced by composition of all the paths with randomly sampled combining coefficients. With the multiple random augmentation processes, the AugMix is able to let the targeted deep network see diverse scenes and enhance the robustness significantly. 
%
%
Note that, existing augmentation methods are primarily for the image classification tasks, which augments the entire image including the background scene and the rain layer (if any) simultaneously and ignores the diversity of rain patterns.
A desired deraining method can not only address different background scenes but also diverse rain patterns.
To this end, we propose a novel data augmentation (\ie \textit{RainMix}) to augment the background scene and rain layer individually, and combine the respective augmentations to achieve the final rainy image. As a result, the targeted deep model can see diverse background scenes as well as various rain patterns.
%

\begin{figure}[t]
\centering
\includegraphics[width=1.0\linewidth]{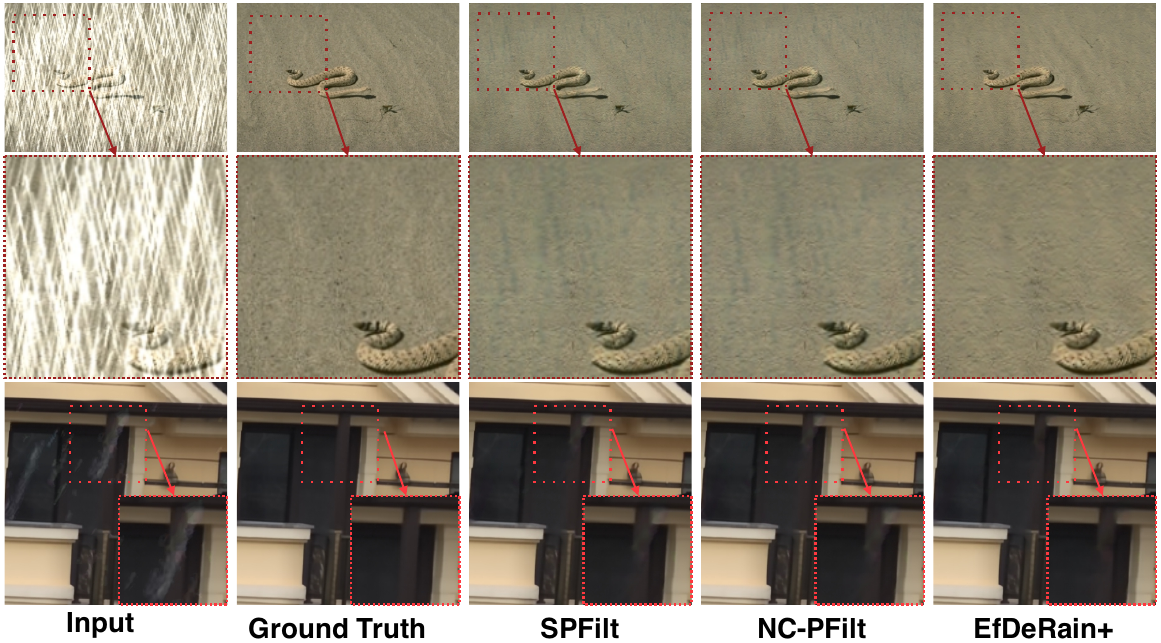}
\vspace{-20pt}
\caption{Visualization of ground truth, and deraining results of spatially-variant predictive filtering (SPFilt) in \secref{sec:img_filter}, naive cascaded predictive filtering (NC-PFilt), and the proposed method in \secref{sec:efderainplus}, respectively. We highlight main differences via red rectangles and arrows.}
\vspace{-15pt}
\label{fig:cfp_vis}
\end{figure}

\section{Deraining as Predictive Filtering}
\label{sec:img_filter}

%
%
We formulate the deraining as a general image filtering problem.
Given a rainy image $\mathbf{I}\in\mathds{R}^{H\times W}$, we conduct the image filtering to remove the rain streaks through
%
\begin{align}\label{eq:imagefilter}
\hat{\mathbf{I}} = \mathbf{K}\circledast\mathbf{I},
\end{align}
%
where $\hat{\mathbf{I}}$ denotes the predicted clean image, and `$\circledast$' is the pixel-wise filtering operation, that is, each pixel of $\mathbf{I}$ is processed by its exclusive kernel and the tensor $\mathbf{K}\in\mathds{R}^{H\times W\times K^2}$ includes the kernels for all pixels.
Specifically, for the $\mathbf{p}$-th position in the image (\ie, $\mathbf{I}[\mathbf{p}]$), we filter it with the $\mathbf{p}$th kernel in $\mathbf{K}$ 
%
\begin{align}\label{eq:pixelfilter}
\hat{\mathbf{I}}[\mathbf{p}] = \sum_{\mathbf{t},\mathbf{q}=\mathbf{p}+\mathbf{t}}\mathbf{K}_{p}[\mathbf{t}]\mathbf{I}[\mathbf{q}],
\end{align}
%
where $\mathbf{K}_p\in\mathds{R}^{K\times K}$ is the matrix reshaped from the $\mathbf{p}$-th vector of $\mathbf{K}$, and $\mathbf{t}$ ranges from $(-\frac{K-1}{2}, -\frac{K-1}{2})$ to $(\frac{K-1}{2}, \frac{K-1}{2})$.
Intuitively, Eq.~\eqref{eq:pixelfilter} linearly combines the $K\times K$ neighboring pixels around $\mathbf{p}$ and use the combination to replace the original $\mathbf{p}$-th pixel. The matrix $\mathbf{K}_p$ determines the weights of neighboring pixels. If the input image is colored, we can filter different channels independently in the same way.

Image filtering is a common image processing operation that has been widely used for image denoising \cite{BakoACMTOG2017}, deblurring \cite{gong2017selfpaced}, sequential interpolation \cite{niklaus2017video}, \etc. 
The basic principle of image filtering is that the clean pixel can be linearly represented or reconstructed by its neighboring pixels. 
We argue that the natural properties of rain streaks, \eg, partial occlusion, transparency, and fog, make the principle also suitable for deraining.
For example, for a rainy pixel (\eg, the center pixel of $9\times 9$ patches in a rainy image \figref{fig:imgfilter}), we can use the combination of its weighted clean neighboring pixels to reconstruct the clean counterpart.
Moreover, \reqref{eq:pixelfilter} is a typical convolution process that has been computationally-accelerated by diverse deep learning platforms, and thus it is possible to realize high-efficiency deraining.

\begin{figure*}[t]
	\centering
	\includegraphics[width=1.0\linewidth]{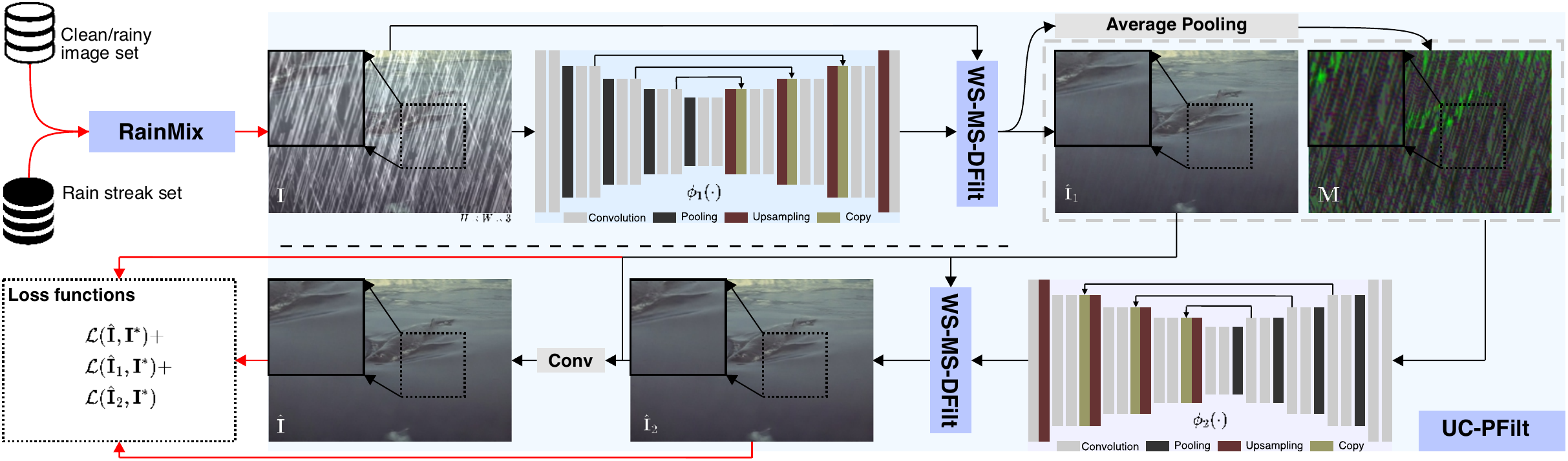}
	\vspace{-20pt}
	\caption{Pipeline of the proposed method covering three main contributions, \ie, uncertainty-aware cascaded predictive filtering (UC-PFilt), weight-sharing multi-scale dilated filtering (WS-MS-DFilt), and a novel data augmentation method RainMix. Note that, the black lines present the process of testing and the red lines indicate the data augmentation and loss functions for training the whole model. }
	\vspace{-15pt}
	\label{fig:archs}
\end{figure*}

To allow effective deraining via filtering, we need to address a key problem, that is, \textit{how to estimate spatially-variant kernels effectively and efficiently.} 
Specifically, rain may cause streak occlusion, fog, and blur, at different positions with dissimilar appearances. 
For example, rain streaks could exhibit different directions, and various transparency levels across the image and are semantically related to the image contents, \eg, the scene depth \cite{hu2019cvpr}. 
As shown in \figref{fig:imgfilter}, we show three different rain patterns within $9\times 9$ patches: in the patch R1, the rain streak partially occludes the object boundary in the background with some transparency. The rain in R2 shows the distinct streak edges with the partial occlusion on a smooth background. In contrast, the patch R3 presents the fog-like rain patterns without obvious streaks.
As a result, the pixel-wise kernels should be adapted to scene information as well as the spatial variations of rain streaks. 
Obviously, hand-craft designed kernels can hardly satisfy such requirements.

%
To allow spatially-variant filtering, we propose to predict the exclusive kernel for each pixel according to the input rainy image itself.
Specifically, we feed the rainy image $\mathbf{I}$ to a CNN $\phi(\cdot)$ denoted as predictive network and estimate the kernels for all pixels by
%
\begin{align}\label{eq:kpn}
\mathbf{K} = \phi(\mathbf{I}).
\end{align}
%
Then, we use the predicted kernels to filter the rainy image $\mathbf{I}$ via \reqref{eq:pixelfilter}. We can simply train the predictive network $\phi(\cdot)$ by minimizing the distance between $\hat{\mathbf{I}}$ and the clean image $\mathbf{I}^*$ via the image quality loss functions, \eg, $L_1$ or SSIM. We denote this method as the spatially-variant predictive filtering (SPFilt) and will detail the implementation in \secref{subsec:basic_impl}. 
Here, we show an example of SPFilt in \figref{fig:imgfilter}:
\begin{itemize}
\item[i.] The naive spatially-variant filtering can effectively remove the rain steak while recovering the occluded boundary, as shown in R1 (See Fig.~\ref{fig:imgfilter} (c)).
\item[ii.] The predicted kernels can adapt to the rain with different strengths. As shown in Fig.~\ref{fig:imgfilter}, from R1 to R3, the rain strength gradually becomes weaker and our method can remove all trace effectively.
Moreover, according to the visualization of predicted kernels, our network can perceive the positions of rain streak. As a result, the predict kernels assign higher weights to non-rainy pixels and lowers ones to rainy pixels, confirming the effectiveness of our method. 
\item[iii.] According to R4, our method does not harm the original boundary and makes it even sharper.
\end{itemize}

Although this simple method has shown great advantages on rain removal, 
we still meet several challenges:
\begin{itemize}
    \item[i.] \textit{how to eliminate residual rain traces.} Although the SPFilt can remove the obvious rain streaks, there are some residual rain traces in the derained images. 
    As shown in \figref{fig:cfp_vis}, compared with the ground truth, the estimation of SPFilt has obvious residual rain traces that mainly locates at the center of rain streaks. 
    Because the neighboring pixels around the center pixel of rain streaks are all corrupted, the reconstruction of the center pixel via \reqref{eq:imagefilter} becomes more difficult. 
    A naive solution is to cascade two predictive filtering to derain the filtering output again.
    However, this straightforward idea is not trivial and a novel method is required. We will detail the challenges in \secref{subsec:uacpfilt} via a naive cascaded solution. 
    
    \item[ii.] \textit{how to achieve effective multi-scale deraining for different scaled rain streaks without harming the efficiency.} In the real world, rain streaks can present diverse scales in the image (\ie, streaks may occupy different pixel areas as the second case in \figref{fig:cfp_vis}). As a result, a multi-scale strategy is required. Nevertheless, it usually requires adding more parameters or recurrence procedure, \eg, wavelet transform \cite{yang2019scalefree}, and recurrent network \cite{li2018recurrent}, which inevitably raises the time or memory costs. How to realize efficient multi-scale deraining is still questionable. 
    
    \item[iii.] \textit{how to train a powerful deraining deep convolution neural network (CNN) to bridge the gap across diverse rain patterns.} Most of the existing deraining DNNs are trained on the synthetic data with limited rain patterns. However, the rain patterns in the real world are diverse and hardly covered by the datasets \cite{yang2020single}. As the second case in \figref{fig:cfp_vis}, SPFilt cannot remove the splash-like rain patterns. Hence, bridging this gap is imperative for real-world applications.

\end{itemize}

To address the above challenges, we first propose \textit{uncertainty-aware cascaded predictive filtering (UC-PFilt)} in \secref{subsec:uacpfilt} that uses the predicted kernels (\ie, $\mathbf{K}$ in \reqref{eq:kpn}) to perceive the difficulties of rain removal on different pixels and train an extra predictive filtering to conduct effective refinement.
Then, we design the \textit{weight-sharing multi-scale dilated filtering (WS-MS-DFilt)} in \secref{subsec:dl_filtering}, which is memory-free and enables slightly extra time cost. Finally, we propose a totally novel data augmentation method, \ie, \textit{RainMix}, in \secref{subsec:rainmix} to bridge the gap across diverse rain patterns. We display the whole pipeline of \ourmethod{} in \figref{fig:archs}.

\begin{figure*}[t]
\centering
\includegraphics[width=1.0\linewidth]{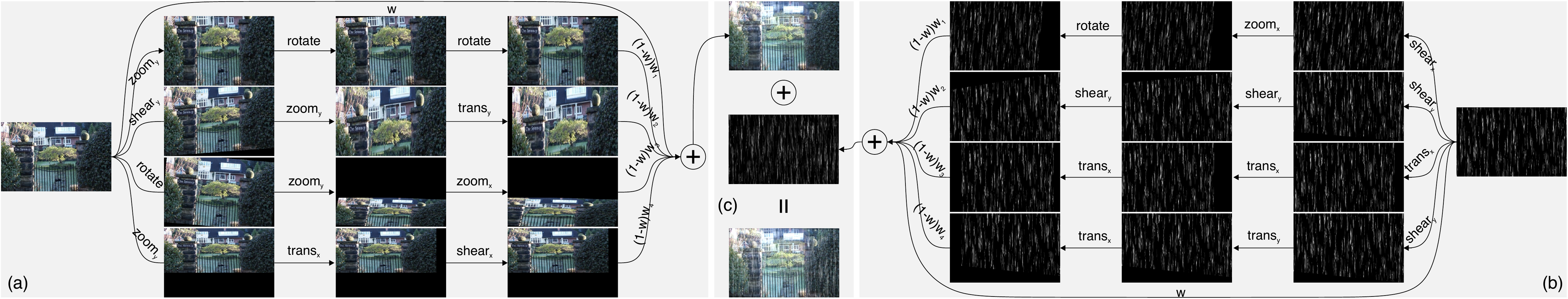}
\vspace{-20pt}
\caption{Pipeline of \textit{RainMix}. (a) and (b) augment a background image and a rainy layer via a series of randomly sampled operations, respectively. Then,  they mix the augmentations and get the final augmented background and rain layer, which synthesizes an rainy image via (c).}
\vspace{-15pt}
\label{fig:rainmix}
\end{figure*}

\section{\ourmethod{}}
\label{sec:efderainplus}

\subsection{Uncertainty-aware Cascaded Predictive Filtering}
\label{subsec:uacpfilt}

%
A straightforward solution for refining the deraining results is to feed the predicted clean image into the predictive filtering again and get the refined image, \ie, 
%
\begin{align}\label{eq:kpn_refine}
\hat{\mathbf{I}} = \hat{\mathbf{K}}\circledast \mathbf{K}\circledast\mathbf{I},~\text{where}~ \mathbf{K} = \phi(\mathbf{I}), \hat{\mathbf{K}} = \phi(\mathbf{K}\circledast\mathbf{I}).
\end{align}
%
However, such a solution fails to enhance the deraining results. As shown in \figref{fig:cfp_vis}, the naive cascaded predictive filtering (NC-PFilt) does not refine the result of SPFilt and its estimation still has obvious artifacts.
To address this issue, we propose \textit{uncertainty-aware cascaded predictive filtering (UC-PFilt)} that has two stages. The first stage filtering generates a primary derained image and an uncertainty map that indicates the difficulties of recovering the pixels via the
predicted kernels in the first stage. 
The second stage filtering is equipped with the uncertainty map and can focus on addressing residual rain traces.
%
%

In \secref{sec:img_filter}, we perform the spatially-variant predictive filtering via \reqref{eq:kpn}, which actually reconstructs rain-free pixels by performing weighted summation on neighboring pixels. 
Note that, the difficulties of reconstructing spatially-variant rainy pixels are different.
Intuitively, it is more difficult to recover a rainy pixel close to the center of a rain streak since its neighboring pixels are also corrupted and the reconstruction might be ambiguous.
If we could identify the difficulties of reconstructing different pixels, we can employ them for further refinement by making the deraining focus on hard pixels.
To this end, we propose to average the weights of the predicted kernels as the difficulty assessors. 
Specifically, for the pixel far from the rain streak center, the pixel is clean and the predicted kernel has high and positive weight at the center and other weights towards zero (See R4 in \figref{fig:imgfilter}), leading to a relative large average across weights. %
When the pixel is close to the center of a rain streak, the estimated kernel assigns negative weights to the rainy neighboring pixels (See R1 in \figref{fig:imgfilter}) and the average of weights becomes small.
The above facts infers that the averages of weights in kernels can assess the difficulties of recovering pixels.
To validate this, we calculate the averages across weights of kernels and see that the pixels far from the rain streaks usually have higher average than the pixels close to the rain streaks (See $\mathbf{M}$ in \figref{fig:archs}).
To utilize this property, we conduct average pooling on each kernel and get an uncertainty score for each pixel. Then, we obtain an uncertainty map denoted as $\mathbf{M}$ that has the same size with the input image (See \figref{fig:archs}).
With this map, given a rainy image $\mathbf{I}$, we can reformulate the cascaded predictive filtering as
%
\begin{align}\label{eq:cas_filtering}
\hat{\mathbf{I}}_2 = \hat{\mathbf{K}}\circledast\hat{\mathbf{I}}_1,~\hat{\mathbf{I}}_1=\mathbf{K}\circledast \mathbf{I},
\end{align}
%
with 
%
\begin{align}\label{eq:cas_net}
\hat{\mathbf{K}} = \phi_2(\hat{\mathbf{I}}_1, \mathbf{M}), \text{and}~\mathbf{K} = \phi_1(\mathbf{I}).
\end{align}
%
where $\phi_1(\cdot)$ and $\phi_2(\cdot)$ are both predictive networks.
The map $\mathbf{M}\in \mathds{R}^{H\times W}$ can very well indicate the rain trace in the rainy image $\mathbf{I}$ and is calculated by performing average pooling on $\mathbf{K}^{H\times W\times K^2}$ in \reqref{eq:kpn} along its third channel. 
Compared with the SPFilt (\ie, \reqref{eq:kpn}), the extra predictive network $\phi_2(\cdot)$ takes $\hat{\mathbf{I}}_1$ and $\mathbf{M}$ as inputs and predicts kernels $\hat{\mathbf{K}}$ for refining $\hat{\mathbf{I}}_1$.
Then, we get the final predicted clean image by fusing the  $\hat{\mathbf{I}}_1$ and $\hat{\mathbf{I}}_2$ via a convolutional layer, \ie,
%
$
\hat{\mathbf{I}} = \text{Conv}(\hat{\mathbf{I}}_1,\hat{\mathbf{I}}_2)
$.
%
We show an example of our method with UC-PFilt in \figref{fig:archs} and \figref{fig:cfp_vis}. Compared with the SPFilt and the naive cascaded predictive filtering (NC-PFilt), the proposed counterpart can remove rain streaks with less artifacts and the final result has almost the same appearance with the rain-free ground truth.

\subsection{Weight-sharing Multi-scale Dilated Filtering} \label{subsec:dl_filtering}
With the image filtering-based deraining in \secref{sec:img_filter}, when the rain streak covers a large region of the image (\ie, large-scale rain streak), the kernel with a large size can use pixels far from the rain for better reconstruction of the clean pixel. 
Following this intuition, we can realize multi-scale deraining straightforwardly by predicting multiple kernels with different sizes for each pixel and getting multiple derained images, and then we fuse these images to produce the final prediction. We formulate this by
%
\begin{align}\label{eq:multi_imgfilter}
\hat{\mathbf{I}}^s = \mathbf{K}^s\circledast\mathbf{I},~s\in\{1,2,\ldots,S\},
\end{align}
%
\begin{align}\label{eq:fusion}
\hat{\mathbf{I}} = \FuncSty{Fusion}(\{\hat{\mathbf{I}}^s|s=1,2,\ldots,S\}),
\end{align}
%
where $\mathbf{K}^s\in\mathds{R}^{H\times W\times (2s+1)^2}$ denotes the kernels under the scale $s$ with its kernel size as $H\times W\times (2s+1)^2$.  
The final deraining result is obtained by fusing $S$ derained images through a fusion function that can be a convolutional layer. 

Although the above solution is able to remove rain streaks having different scales, it requires heavy memory and time costs because larger kernels greatly aggravate the computation costs. Specifically, with the kernel size $K^s=2s+1$, the complexity of performing pixel-wise filtering is $\mathcal{O}(HW({2s+1})^2)$ and we have the complexity $\mathcal{O}(HW\sum_{s=1}^S({2s+1})^2)$ for performing $S$-scale deraining. 

For efficient multi-scale deraining, we propose the weight-sharing multi-scale dilated filtering. Specifically, after the kernel prediction via \reqref{eq:kpn}, we get $\mathbf{K}$ and dilate it to $S$ variants by interpolating empty positions between the elements in $\mathbf{K}$
%
\begin{align}\label{eq:dilated_filter}
&\mathbf{K}^s[\mathbf{t}] = \left\{
\begin{array}{ll}
& \mathbf{K}[\frac{\mathbf{t}-1}{2}], ~\mathbf{t} \bmod 2 \neq (0,0)\\
& \text{Empty},~\mathbf{t} \bmod 2 = (0,0)
\end{array}
\right., \\
&\mathbf{t}=(-\frac{K^s-1}{2}, -\frac{K^s-1}{2}) ,\ldots,(\frac{K^s-1}{2}, \frac{K^s-1}{2}). \nonumber
\end{align}
%
With \reqref{eq:dilated_filter}, the multi-scale kernels share the same weights predicted from \reqref{eq:kpn}. Inspired by the dilated convolution \cite{Yu2016iclr}, we can further simplify the multi-scale filtering (\ie, \reqref{eq:multi_imgfilter}) without explicitly calculating $\{\mathbf{K}^s\}$, \ie,
%
\begin{align}\label{eq:dilated_pixelfilter}
\hat{\mathbf{I}}^s[\mathbf{p}] = \sum_{\mathbf{t},\mathbf{q}=\mathbf{p}+s\mathbf{t}}\mathbf{K}_{p}[\mathbf{t}]\mathbf{I}[\mathbf{q}].
\end{align}
%
Then, the complexity of multi-scale filtering reduces from $\mathcal{O}(HW(\sum_{s=1}^S({2s+1})^2))$ to $\mathcal{O}(9HW)$ without extra memory cost.
Then, we can update \reqref{eq:cas_filtering} by replacing the naive filtering `$\circledast$' with our WS-MS-DFilt.
Furthermore, we will conduct detailed analysis in \secref{subsec:multiscale_variants}, which demonstrates that weight-sharing enables more effective deraining than a naive strategy predicting multi-scale filters independently.

\subsection{\textit{RainMix} for Bridging the Gap to Diverse Rains}
\label{subsec:rainmix}
How to reduce the gap of rainy images in the training set to diverse rain patterns in the real world is still an open problem. 
We address this challenge from the angle of data augmentation and propose a novel augmentation method for training powerful deraining models. 
The basic principle behind this idea is to make the desired CNN see enough and diverse rain patterns as well as background scenes that possibly can happen in the real world. As a result, the CNN can address similar situations when we test it with real-world data. 
Existing data augmentation methods \cite{yun2019cutmix,hendrycks2020augmix} augment images by conducting image transformations, \eg, rotation, zooming, translation, \etc. 
Such solution changes the rain streaks and background simultaneously and cannot enrich the rain patterns. 
For example, when we translate a rainy image, both streaks and background are shifted. As a result, the rain patterns are not changes.
To alleviate this issue, we propose to augment the rain layer and background, respectively, and use their augmented counterparts to synthesize new rainy images.
%
%

\begin{figure*}[t]
	\centering
	\includegraphics[width=1.0\linewidth]{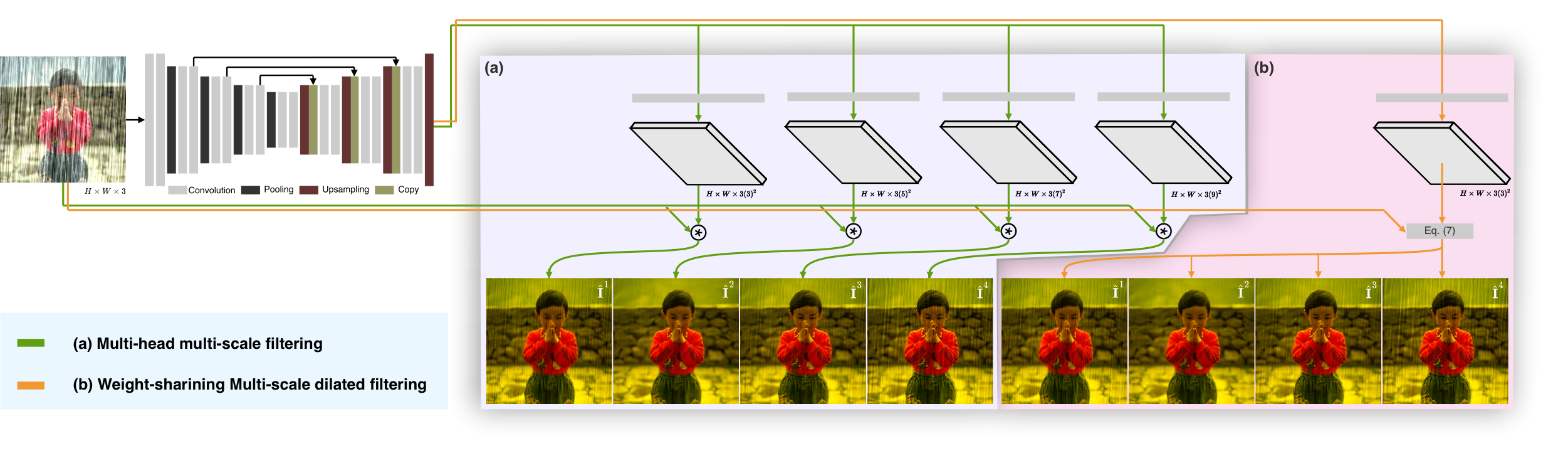}
	\vspace{-30pt}
	\caption{Pipelines of the muitli-head multi-scale filterin (MH-MS-Filt) and the proposed weight-sharing multi-scale dilated filtering (WS-MS-DFilt). We also visualize the deraining results of ground truth, single-predictive network, and cascaded-predictive network, respectively.}
	\vspace{-15pt}
	\label{fig:multi-scale}
\end{figure*}

To this end, we focus on two key issues when designing the augmentation method, \ie, how to attain realistic rain layers and how to make the augmented rainy scenes diverse enough.
For the first issue, we identify two ways to get a set of rain layers (\ie, $\mathcal{R}$). First, given a training dataset, we can collect rain layers by subtracting the clean images from the corresponding rainy images. Second, we borrow the rain streaks generated by \cite{garg2006siggraph} that constructs a dataset of real rain streaks by considering different lighting and viewing conditions. 
Then, we can augment these rain layers via transformations to simulate the influences of various natural factors, \ie, the wind, the light reflection, and refraction, \etc, which may happen in the real world.
To generate diverse rainy scenes (\ie, the second issue), given a background and a rain layer, we propose to conduct multiple random sampling processes to augment the background and the rain layer, respectively, and obtain a series of augmentations.
Then, we can mix augmentations up and add the augmented rain layer to the augmented background to simulate the complex rain patterns and scene variations in the real world. For intuitive understanding, we name our augmentation method as \textit{RainMix} and present the \textit{RainMix}-based learning algorithm in Algorithm~\ref{alg} and \figref{fig:rainmix}.

Specifically, at each training iteration, we load a background that can be a clean image or a rainy image (See line 15 in Algorithm~\ref{alg}), and randomly sample a rain layer from $\mathcal{R}$. 
Then, we can generate an augmented background and an augmented rain layer via \textit{RainMix}, respectively (\ie, line 16 in Algorithm~\ref{alg}). After that, we add the augmented rain layer to the augmented background  (See line 17), thus we obtain a new rainy image for training $\phi_1(\cdot)$, $\phi_2(\cdot)$, and \FuncSty{Fusion} in \reqref{eq:fusion}. 
Regarding the details of \textit{RainMix}, given a tensor $\mathbf{X}^\text{org}$ that could be a background and a rain layer (\ie, line 2), \textit{RainMix} performs a series of transformations on $\mathbf{X}^\text{org}$ via randomly sampled and combined $M$ operations (\ie, line 6-8). We repeat this process for $N$ times and get $N$ transformed inputs that are further mixed via the weights from Dirichlet distribution (See line 9) and are further blended with the original sampled rain map via the weight from Beta distribution (\ie, line 4 and 11). Intuitively, the multiple random processes simulate the diverse rain patterns and scene variances in the real world. We give an example of \textit{RainMix} for generating a rainy image in Figure~\ref{fig:rainmix}.

\begin{algorithm}[tb]
	{
		\caption{\small{Learning Deraining via \textit{RainMix}}}\label{alg}
		\KwIn{$\phi_1(\cdot)$, $\phi_2(\cdot)$, $\FuncSty{Fusion}(\cdot)$, Loss function $\mathcal{L}$, Rainy Images $\mathcal{I}$, Clean Images $\mathcal{I}^*$, rain layer set $\mathcal{R}$, Augmentation operation set $\mathcal{O}=\{\text{rot},\text{shear}_{x/y},\text{trans}_{x/y}, \text{zoom}_{x/y}\}$.}
		\KwOut{$\phi_1(\cdot)$, $\phi_2(\cdot)$, and $\FuncSty{Fusion}(\cdot)$.}
        \SetKwFunction{FMain}{RainMix}
        \SetKwProg{Fn}{Function}{:}{}
        \Fn{\FMain{$\mathcal{X}$}}{
        Sample a tensor $\mathbf{X}^\text{org}\sim\mathcal{X}$\;
        Initialize an empty tensor $\mathbf{X}^\text{mix}$\;
        Sample mixing weights $(w_1,\ldots, w_{N})\sim\text{Dirichlet}$\;
        \For{$i=1\ \mathrm{to}\ N$}{
            Sample $M$ operations $(\text{o}_1,\ldots,\text{o}_M)\sim\mathcal{O}$\;
            Combine via $\text{o}_{12}=\text{o}_2\text{o}_1$ and $\text{o}_{12\ldots M}=\text{o}_M\ldots\text{o}_2\text{o}_1$\;
            Sample $\text{o}\sim\{\text{o}_{1},\text{o}_{12},\text{o}_{123},\ldots,\text{o}_{12\ldots M}\}$\;
            $\mathbf{X}^\text{mix}+=w_i\text{o}(\mathbf{X}^\text{org})$
            }
        Sample a blending weight $w\sim\text{Beta}$\;
        \textbf{return} $\mathbf{X}=w\mathbf{X}^\text{org}+(1-w)\mathbf{X}^\text{mix}$\;
        }
        \textbf{End function}\;
 		\For{$i=1\ \mathrm{to}\ |\mathcal{I}|$}{
 		    Sample an image pair via $(\mathbf{I},\mathbf{I}^*)\sim(\mathcal{I},\mathcal{I}^*)$\; 		    
 		    Augment a rain layer and a bacground via $\hat{\mathbf{R}}=\FuncSty{RainMix}(\mathcal{R})$ and $\hat{\mathbf{I}}=\FuncSty{RainMix}(\{\mathbf{I},\mathbf{I}^*\})$\;
            Generate a rainy image via $\hat{\mathbf{I}}'=\hat{\mathbf{R}}+\hat{\mathbf{I}}$\;
            Predict kernels via \reqref{eq:cas_net}
            and Derain via \reqref{eq:cas_filtering} by taking $\hat{\mathbf{I}}'$ as the input rainy image\;
            Calculate loss function and do back-propagation\;
            Update parameters of $\phi_1(\cdot)$, $\phi_2(\cdot)$, and $\FuncSty{Fusion}(\cdot)$\;
    	}
	}
\end{algorithm}

\section{Discussion and Implementations}
\label{sec:analysis}
In this section, we first introduce the basic implementations of our spatially-variant predictive filtering (SPFilt) and uncertainty-aware cascaded predictive filtering (UC-PFilt) in \secref{subsec:basic_impl}. Then, we discuss the architectural variants of SPFilt and UC-PFilt in \secref{subsec:arch_variants}. In \secref{subsec:multiscale_variants}, we study and compare two multi-scale deraining strategies, \ie, the proposed \textit{weight-sharing multi-scale dilated filtering (WS-MS-DFilt)} and a naive multi-head multi-scale strategy. 
Then, we validate the effectiveness of \textit{RainMix} in \secref{subsec:aug_variants}.
Note that, the studies in \secref{subsec:multiscale_variants} and \secref{subsec:aug_variants} are based on the basic implementation of SPFilt.
Finally, we equip UC-PFilt with the multi-scale strategy and \textit{RainMix} and introduce our final implementation in \secref{subsec:final_impl}.

%
\begin{table*}[t]
	\centering
	\caption{Architecture variants of predictive network (\ie, $\phi(\cdot)$ in \reqref{eq:kpn} and $\phi_1(\cdot)$ and $\phi_2(\cdot)$ in \reqref{eq:cas_net}) with different depths. The `[$\cdot$]' denotes the concatenation operation. Each convolution layer is followed by a batch normalization layer and an activation layer.}\label{tab:arch}
	\vspace{-5pt}
	\begin{tabular}{l|c|c|l|ll|ll|ll}
		\toprule
		& input & output & output size  & \multicolumn{2}{c|}{49-layer} & \multicolumn{2}{c|}{33-layer} & \multicolumn{2}{c}{17-layer} \\\midrule
		Block1  & $\mathbf{I}$ & $\mathbf{x}_1$ & $256\times256$      & conv{[}$3\times3$, 64{]}                 & $\times3$                & conv{[}$3\times3$, 64{]}                 & $\times2$                & conv{[}$3\times3$, 64{]}                 & $\times1$                \\
		Block2  & $\mathbf{x}_1$ & $\mathbf{x}_2$ & $128\times128$      & conv{[}$3\times3$, 128{]}                & $\times3$                & conv{[}$3\times3$, 128{]}                & $\times2$                & conv{[}$3\times3$, 128{]}                & $\times1$                \\
		Block3  & $\mathbf{x}_2$ & $\mathbf{x}_3$ & $64\times64$        & conv{[}$3\times3$, 256{]}                & $\times3$                & conv{[}$3\times3$, 256{]}                & $\times2$                & conv{[}$3\times3$, 256{]}                & $\times1$                \\
		Block4  & $\mathbf{x}_3$ & $\mathbf{x}_4$ & $32\times32$        & conv{[}$3\times3$, 512{]}                & $\times3$                & conv{[}$3\times3$, 512{]}                & $\times2$                & conv{[}$3\times3$, 512{]}                & $\times1$                \\
		Block5  & $\mathbf{x}_4$ & $\mathbf{x}_5$ & $16\times16$        & conv{[}$3\times3$, 512{]}                & $\times3$                & conv{[}$3\times3$, 512{]}                & $\times2$                & conv{[}$3\times3$, 512{]}                & $\times1$                \\
		Block6  & $\mathbf{x}_5$ & $\mathbf{x}_6$ & $32\times32$        & conv{[}$3\times3$, 512{]}                & $\times3$                & conv{[}$3\times3$, 512{]}                & $\times2$                & conv{[}$3\times3$, 512{]}                & $\times1$                \\
		Block7  & [$\mathbf{x}_6$,$\mathbf{x}_4$] & $\mathbf{x}_7$ & $64\times64$        & conv{[}$3\times3$, 256{]}                & $\times3$                & conv{[}$3\times3$, 256{]}                & $\times2$                & conv{[}$3\times3$, 256{]}                & $\times1$                \\
		Block8  & [$\mathbf{x}_7$,$\mathbf{x}_3$] & $\mathbf{x}_8$ & $128\times128$      & conv{[}$3\times3$, 27{]}                 & $\times3$                & conv{[}$3\times3$, 27{]}                 & $\times2$                & conv{[}$3\times3$, 27{]}                 & $\times1$                \\
		Conv\_core & [$\mathbf{x}_8$,$\mathbf{x}_2$] & $\mathbf{K}$ & $256\times256$      & conv{[}$1\times1$, 27{]}                 & $\times1$                & conv{[}$1\times1$, 27{]}                 & $\times1$                & conv{[}$1\times1$, 27{]}                 & $\times1$                \\ 
		\bottomrule
	\end{tabular}
	\vspace{-5pt}
\end{table*}
%

\subsection{Basic Implementation}
\label{subsec:basic_impl}
To realize deraining via SPFilt and UC-PFilt, we need to train $\phi(\cdot)$ in \reqref{eq:kpn} for SPFilt and $\phi_1(\cdot)$ and $\phi_2(\cdot)$ in \reqref{eq:cas_net} for UC-PFilt. We detail the architecture setups, loss functions, and training details in the following.

{\bf Architectures for predictive networks.} We borrow UNet \cite{ronneberger2015u} as our basic architecture. It takes a colorful colored rainy image $\mathbf{I}$ as the input whose size is $256\times 256\times 3$. To let the output be the pixel-wise kernels (\ie, $\mathbf{K}$), we set the output size as $256\times 256 \times 27$ that is reshaped as $256\times 256 \times 3\times 9$ that assigns each pixel a kernel at each channel. We detail the architecture in Table~\ref{tab:arch}. We can set $\phi(\cdot)$, $\phi_1(\cdot)$, and $\phi_2(\cdot)$ as the networks with different depths in Table~\ref{tab:arch}. In \secref{subsec:arch_variants}, we discuss the influence of different architectural variants.

{\bf Loss function for SPFilt.}
We employ two loss functions to train the predictive network, \ie, the $L_1$ and SSIM loss functions \cite{zhao2016loss}. Given a derained image, \ie, $\hat{\mathbf{I}}$, and the corresponding clean image $\mathbf{I}^{*}$ as the ground truth, we have
%
\begin{align}\label{eq:loss}
\mathcal{L}(\hat{\mathbf{I}}, \mathbf{I}^*) = \|\hat{\mathbf{I}}-\mathbf{I}^*\|_1-\lambda~\text{SSIM}(\hat{\mathbf{I}}, \mathbf{I}^*)
\end{align}
%
where we fix $\lambda=0.2$ for all experiments.

{\bf Loss function for UC-PFilt.} UC-PFilt predicts three rain-free images, \ie, $\hat{\mathbf{I}}$, $\hat{\mathbf{I}}_1$, and $\hat{\mathbf{I}}_2$, and we can apply the loss function \reqref{eq:loss} to the three images, respectively,
%
\begin{align}\label{eq:dualloss}
\mathcal{L}_\text{uc}(\hat{\mathbf{I}}, \hat{\mathbf{I}}_1, \hat{\mathbf{I}}_2, \mathbf{I}^*) = \mathcal{L}(\hat{\mathbf{I}}, \mathbf{I}^*)+\mathcal{L}(\hat{\mathbf{I}}_1, \mathbf{I}^*)+\mathcal{L}(\hat{\mathbf{I}}_2, \mathbf{I}^*),
\end{align}
%

{\bf Training details for SPFilt.} In this basic implementation, we train $\varphi(\cdot)$ without any data augmentation strategies. We calculate the losses with derained images and their clean counterparts provided by the training dataset. During the training iteration, we fix the learning rate as $1e^{-4}$ and terminate the training when the loss stop falling. We will validate the proposed \textit{RainMix} based on this training setup.

{\bf Training details for UC-PFilt.}
To allow effective training, we first train $\phi_1(\cdot)$ in the same way of training the SPFilt with \reqref{eq:loss} and then jointly train the parameters of $\phi_1(\cdot)$, $\phi_2(\cdot)$, and the fusion convolutional layer via \reqref{eq:dualloss}.
Under the basic implementation, we do not perform any data augmentation and do not equip the multi-scale module.

In the following analysis experiments, we employ the Rian100H dataset \cite{yang2017cvpr,yang2019joint} for training and testing. We adopt the peak signal-to-noise ratio (PSNR) and structural similarity index (SSIM) as the restoration evaluation \cite{zhao2016loss} and use Giga floating-point operations per second (GFLOPS) for complexity comparison.

\subsection{Architectural Variants of Predictive Filtering}
\label{subsec:arch_variants}
We discuss the variants of the predictive networks $\phi(\cdot)$ in \reqref{eq:kpn} and $\phi_1(\cdot)$ and $\phi_2(\cdot)$ in \reqref{eq:cas_net} based on the UNet \cite{ronneberger2015u} by setting different depths.

{\bf Variants of SPFilt.} We construct three predictive networks by setting the number of convolutional layers in the blocks as three, two, and one, respectively. As shown in the \tableref{tab:arch}, we acquire three variants of $\phi(\cdot)$ with 49~layers, 33~layers, and 17~layers, respectively.
As a common understanding, we observe that deeper predictive network (\ie $\phi(\cdot)$) achieves higher PSNR and SSIM in \tableref{tab:arch_dual}. For example, the 49-layer and 33-layer networks achieve the highest and second highest PSNR and SSIM, respectively.

{\bf Variants of UC-PFilt.}
For UC-PFilt implementation, a key problem is how to set the depths for the two predictive networks. 
Intuitively, we can set $\phi_1(\cdot)$ and $\phi_2(\cdot)$ as the deepest architecture, \ie, the 49-layer network in \tableref{tab:arch}.
However, such a setup would increase the time cost and it is more difficult to train.
We extensively study the influence of the UC-PFilt's architecture by considering different depths of $\phi_1(\cdot)$ and $\phi_2(\cdot)$.
Specifically, we first train SPFilts with three depths (\ie, 49, 33, and 17 layers) as the $\phi_1(\cdot)$, respectively. Then, we equip each $\phi_1(\cdot)$ with $\phi_2(\cdot)$ that also has three different depths (\ie, 49, 33, and 17 layers). As a result, we have three SPFilts and nine UC-PFilts. 
We evaluate these variants on the Rain100H dataset with PSNR, SSIM, GFLOPS, and time costs as the metrics. 
As shown in \tableref{tab:arch_dual}, we have the following observations: \textit{First}, compared with SPFilt, the UC-PFilts always achieve higher PSNR and SSIM, which demonstrates that the extra predictive network $\phi_2(\cdot)$ does further enhance the image quality.
\textit{Second}, the deeper $\phi_2(\cdot)$ does not lead to higher quality enhancement. For example, when $\phi_1(\cdot)$ is a 49-layer network, the UC-PFilt with the 17-layer $\phi_2(\cdot)$ achieves higher SSIM than the ones with 49-layer and 33-layer $\phi_2(\cdot)$. 
One possible reason is that deeper $\phi_1(\cdot)$ and $\phi_2(\cdot)$ are more difficult to train. 
In particular, UC-PFilt with 49-layer $\phi_1(\cdot)$ and 17-layer $\phi_2(\cdot)$ achieve the highest PSNR and SSIM.

%
\begin{table*}[t]
\centering
\caption{Evaluation results of nine UC-PFilts and three SPFilts with different depths of $\phi_1(\cdot)$ and $\phi_2(\cdot)$. When $\phi_2(\cdot)$ is empty, we mean the variants as the single-predictive networks.
We highlight the top results with bold fonts.
}\label{tab:arch_dual}
	\vspace{-5pt}
\begin{tabular}{l|llll|llll|llll|llll}
\toprule
 & \multicolumn{4}{c|}{PSNR} & \multicolumn{4}{c|}{SSIM} & \multicolumn{4}{c|}{GFLOPS} &  \multicolumn{4}{c}{Time (ms)} \\
\midrule
\diagbox[width=3.5em,trim=l]{$\phi$/$\phi_1$}{$\phi_2$} 
   & -     & 49    & 33    & 17     & -      & 49     & 33     & 17             & -     & 49    & 33    & 17    & -    & 49   & 33   & 17 \\
\midrule
49 & 31.39 & 32.51 & 32.51 & \first{32.52}  & 0.9199 & 0.9296 & 0.9292 & \first{0.9303} & 37.12 & 74.23 & 62.57 & 50.91 & 4.14 & 7.82 & 7.16 & 6.19 \\
33 & 31.10 & 31.49 & 31.87 & 31.74          & 0.9115 & 0.9193 & 0.9245 & 0.9236         & 25.46 & 62.57 & 50.91 & 39.25 & 3.22 & 7.01 & 6.09 & 5.30 \\
17 & 30.77 & 31.42 & 30.80 & 31.23          & 0.9048 & 0.9185 & 0.9107 & 0.9183         & 13.80 & 50.91 & 39.25 & 27.59 & 2.88 & 6.04  & 5.23 & 4.30 \\
\bottomrule
\end{tabular}
	\vspace{-10pt}
\end{table*}
%

\subsection{Multi-scale Dilated Filtering Variants}
\label{subsec:multiscale_variants}
In this part, we study the effectiveness and efficiency of the proposed weight-sharing multi-scale dilated filtering (WS-MS-DFilt). We start with a naive multi-scale strategy, that is, we use the predictive network $\phi(\cdot)$ to estimate multi-scale kernels through multiple predictive heads independently and perform multi-scale deraining via \reqref{eq:multi_imgfilter}. 
To this end, we extend the last layer of the architecture of $\phi(\cdot)$ in \secref{subsec:basic_impl} to $S$ convolutional layers where $S$ is the number of scales in \reqref{eq:multi_imgfilter}. Each convolutional layer outputs a tensor $\mathbf{K}^s$ with size of $H\times W\times 3(2s+1)^2$, which is used to predict the clean image at scale $s$ (\ie, $\mathbf{I}^s$ in \reqref{eq:multi_imgfilter}).
We show the architecture in \figref{fig:multi-scale}~(a) and denote it as the multi-head multi-scale filtering (MH-MS-Filt).
Then, we implement the proposed multi-scale strategy in \secref{subsec:dl_filtering} via \reqref{eq:dilated_pixelfilter} with the pipeline shown in \figref{fig:multi-scale}~(b).
The Above two solutions fuse the predicted multi-scale images via \reqref{eq:fusion} and output the final prediction.

We compare the two solutions on Rain100H dataset with different scales, \ie, $S\in\{2,3,4\}$. For example, when $S=2$, we predict $\hat{\mathbf{I}}^1$ and $\hat{\mathbf{I}}^2$ and fuse them via a $3\times 3$ convolutional layer as the $\FuncSty{Fusion}(\cdot)$ in \reqref{eq:fusion}. All results are reported in \tableref{tab:multi-scale} and we have the following observations: \textit{First}, WS-MS-DFilt outperforms the naive multi-scale strategy, \ie, MH-MS-Filt, across all scale numbers. \textit{Second}, MH-MS-Filt's GFLOPS and time costs gradually increase as the $S$ becomes larger. In contrast, WS-MS-DFilt has similar GFLOPS and time costs across different scales with much lower time costs than MH-MS-Filt with $S=4$. 
\textit{Third}, compared with the single-scale predictive filtering (\ie, the $\phi_1$ with the 49-layer network and empty $\phi_2$ in \tableref{tab:arch_dual}), WS-MS-DFilt with $S=4$ achieves much higher SSIM (\ie, 0.9023 vs. 0.8989) but but with similar GFLOPS and time costs. 

\begin{table*}[t]
\centering
\caption{Comparison results of the multi-head multi-scale filtering (MH-MS-Filt) and the proposed weight-sharing multi-scale dilated filtering (WS-MS-DFilt) on Rain100H under three scale numbers. We highlight the best and second best results with \first{red} and \second{green} fonts.}\label{tab:multi-scale}
	\vspace{-5pt}
\begin{tabular}{l|ll|ll|ll|ll}
\toprule
 & \multicolumn{2}{c|}{PSNR} & \multicolumn{2}{c|}{SSIM} & \multicolumn{2}{c|}{GFLOPS} & \multicolumn{2}{c}{Time (ms)} \\
\midrule
 & MH-MS-Filt & WS-MS-DFilt & MH-MS-Filt & WS-MS-DFilt & MH-MS-Filt & WS-MS-DFilt & MH-MS-Filt & WS-MS-DFilt \\
\midrule
$S=2$ & 29.61 & 30.17 & 0.9020 & 0.9054 & 43.06 & 37.13 & 4.78 & 4.11 \\
$S=3$ & 29.29 & \second{31.14} & 0.9048 & \second{0.9182} & 63.71 & 37.14 & 5.94 & 4.04 \\
$S=4$ & 29.97 & \first{31.47} & 0.9091 & \first{0.9210} & 124.01 & 37.15 & 7.35 & 4.14 \\
\bottomrule
\end{tabular}
	\vspace{-5pt}
\end{table*}
%

\subsection{Augmentation Variants for \textit{RainMix}}
\label{subsec:aug_variants}

We argue that the main effectiveness of \textit{RainMix} stems from the high diversity introduced by the multiple random sampling processes, and we study the \textit{RainMix} from two aspects: how does the number of mixed rain layers or backgrounds and the number of random sampled operations (\ie, $N$ and $M$ in Algorithm~\ref{alg}, respectively) affect the deraining training? 
To this end, we use \textit{RainMix}s with different $N$ and $M$ to train the 49-layer predictive network based on the basic implementation introduced in \secref{subsec:basic_impl}.

As reported in \tableref{tab:rainmix_var}, we see that: the PSNR and SSIM values after deraining gradually increase as the number of mixed rain layers and backgrounds (\ie, $N$) becomes larger and tend to be smaller as the number of random sampled operations (\ie, $M$) is smaller. 
This is because the augmented rain scenes contain more diverse rain patterns when $N$ and $M$ become larger which allows the targeted model see more different rainy images.
%

\begin{table}[t]
\centering
\setlength\tabcolsep{4pt}
\caption{Comparison results of \textit{RainMix}s with different $N$ (\ie, number of mixed rain layers) and $M$ (\ie, number of random sampled operations) in Algorithm~\ref{alg}. We use the \textit{RainMix}s to train the basic implementation (\secref{subsec:basic_impl}) with a 49-layer predictive network on Rain100H. We highlight the best results with bold fonts.
}\label{tab:rainmix_var}
	\vspace{-5pt}
\begin{tabular}{l|lllllll}
\toprule
$(N,M)$ & - & (1,3) & (2,3) & (3,3) & (4,3) & (4,2) & (4,1) \\
\midrule
PSNR & 31.39  & 32.58 & 32.66 & 32.66 & \first{32.69} & 32.65 & 32.65 \\
SSIM & 0.9199 & 0.9241 & 0.9255 & 0.9255 & \first{0.9271} & 0.9256 & 0.9254 \\
\bottomrule
\end{tabular}
	\vspace{-10pt}
\end{table}

\begin{figure*}[t]
\centering
\includegraphics[width=1.0\linewidth]{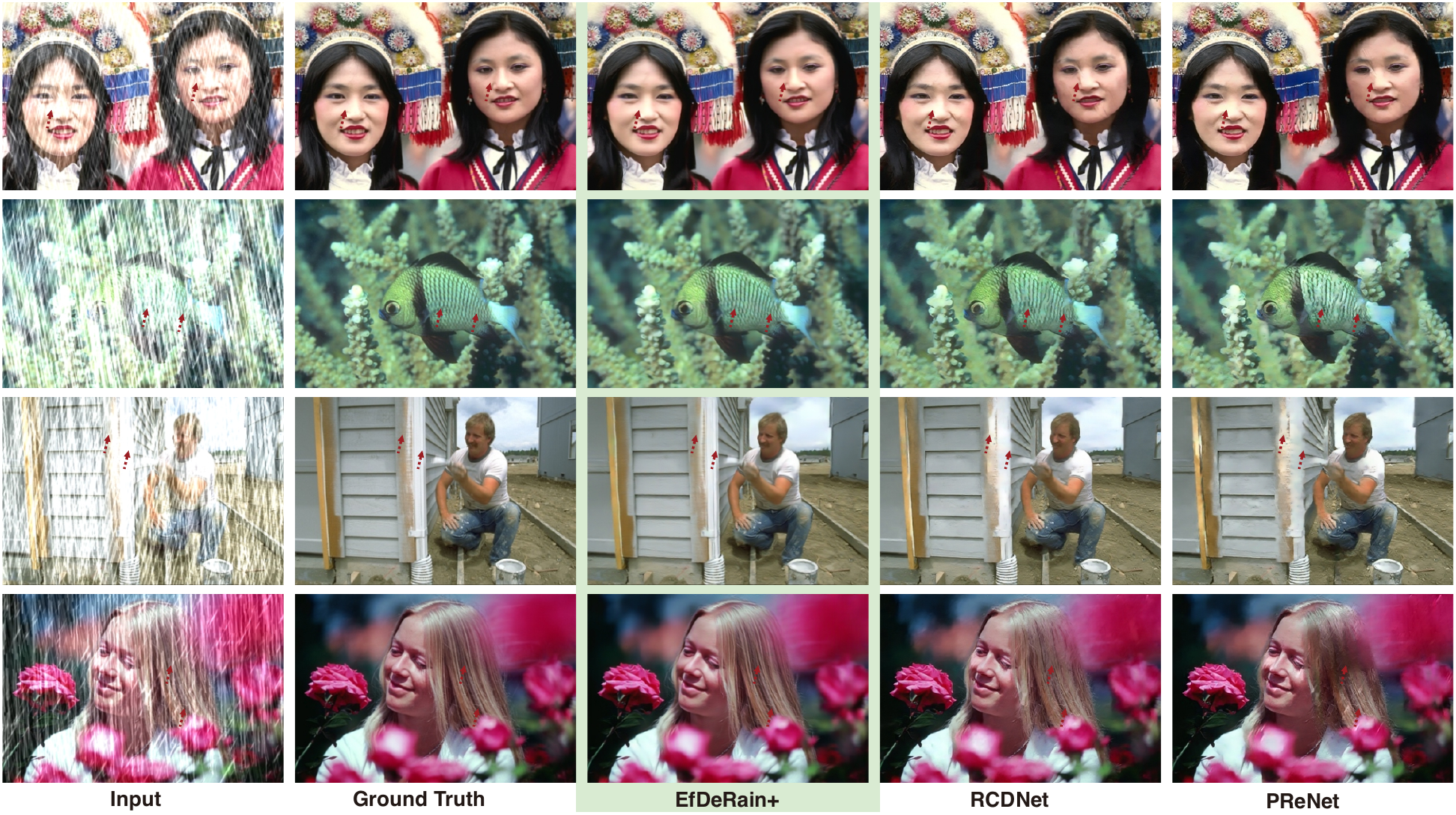}
\vspace{-20pt}
\caption{Visualization comparison with two state-of-the-art methods, \ie, RCDNet \cite{wang2020a} and PReNet \cite{ren2019cvpr}, on four cases from Rain100H.  }
\vspace{-10pt}
\label{fig:rain100h}
\end{figure*}

\subsection{Final implementation}
\label{subsec:final_impl}

The above discussions have demonstrated the effectiveness of the proposed weight-sharing multi-scale dilated filtering, \textit{RainMix} for training SPFilt and UC-PFilt, respectively. Here, we combine all these modules and obtain the final implementation of our method. Specifically, we equip the $\phi_1(\cdot)$ and $\phi_2(\cdot)$ of UC-PFilt with the proposed multi-scale module (\ie, \reqref{eq:dilated_filter} and \reqref{eq:fusion}), and train the two networks $\phi_1(\cdot)$ and $\phi_2(\cdot)$ with proposed \textit{RainMix}.
In terms of the architectures of $\phi_1(\cdot)$ and $\phi_2(\cdot)$, we select 49-layer $\phi_1(\cdot)$ and 17-layer $\phi_2(\cdot)$ for UC-PFilt since such a variant achieves the highest SSIM with moderate time cost (\ie, 6.19~ms) and GFLOPS according to the experimental results in \tableref{tab:arch_dual} and discussions in \secref{subsec:arch_variants}.
In terms of the training details, we follow the setups for UC-PFilt introduced in \secref{subsec:basic_impl}. 
We will validate each module via the ablation study in \secref{subsec:ablation_study}.

\section{Experimental Results}
\label{sec:experiment}

\begin{table*}[t]
	\centering
    \setlength\tabcolsep{1pt}
	\caption{Experimental results on Rain100H dataset \cite{yang2017cvpr}.
    We highlight the top results with bold fonts.
	}
	\label{tab:rain100h}
		\vspace{-5pt}
	\begin{tabular}{l|lllllllllll}
		\toprule
		 &  Clear \cite{fu2017tip}  & DDN \cite{fu2017removing} & RESCAN \cite{li2018recurrent} & SIRR \cite{wei2019cvpr} & PReNet \cite{ren2019cvpr} & SPANet \cite{wang2019spatial} & JORDERE \cite{yang2019joint} & CVID \cite{cvid} & RCDNet \cite{wang2020a} & D-DAM \cite{zhang2021dual} & \ourmethod{} \\ 
		 \midrule
		PSNR &  15.33  &   22.85    &   29.62     &    22.47   &   30.11    &  25.11  &  30.50  & 27.93  & \second{31.28}  & 30.35 & \first{34.57} \\ 
		SSIM &  0.7421 &   0.7250   &   0.8720    &    0.7164  &   0.9053   &  0.8332 &  0.8976 & 0.8765 & \second{0.9093} & 0.907 & \first{0.9513}\\
		Time~(ms) &  133.7  &   67.6     &   21.25     &    363.9   &   45.4     &  \second{10.4}   &  12.2  & 272.2  & 468.7  & - & \first{6.3} \\
		\bottomrule
	\end{tabular}
		\vspace{-5pt}
\end{table*}

%
\subsection{Setups}
\label{subsec:setup}
We conduct the single-image and video deraining to validate the effectiveness of our method.

{\bf Datasets.} To comprehensively validate and evaluate our method, we conduct the comparison and analysis experiments on four popular datasets for single-image deraining, including \emph{Rain100H} \cite{yang2017cvpr}, \emph{Rain1400} \cite{fu2017removing} synthetic datasets, the recently proposed \emph{SPA} real rain dataset \cite{wang2019spatial}, and the real \emph{Raindrop} dataset \cite{qian2018cvpr}. We select \emph{Rain100H} since it contains five types of rain streaks and is much more challenging than other synthesized datasets, \eg, Rain100L \cite{yang2017cvpr}, Rain12 \cite{li2016rain}, and Rain800 \cite{zhang2018iccv}. Rain1400 \cite{fu2017removing} contains 9,100 and 4,900 training and testing images and is a much larger dataset. 
In terms of the video deraining, we use NTURain dataset \cite{chen2018robust} for evaluation and comparison. 

{\bf Metrics.} We employ the commonly used peak signal to noise ratio (PSNR) and structural similarity (SSIM) as the quantitative evaluation metric for all datasets. In general, the larger PSNR and SSIM indicate better deraining results.

\begin{figure*}[t]
\centering
\includegraphics[width=1.0\linewidth]{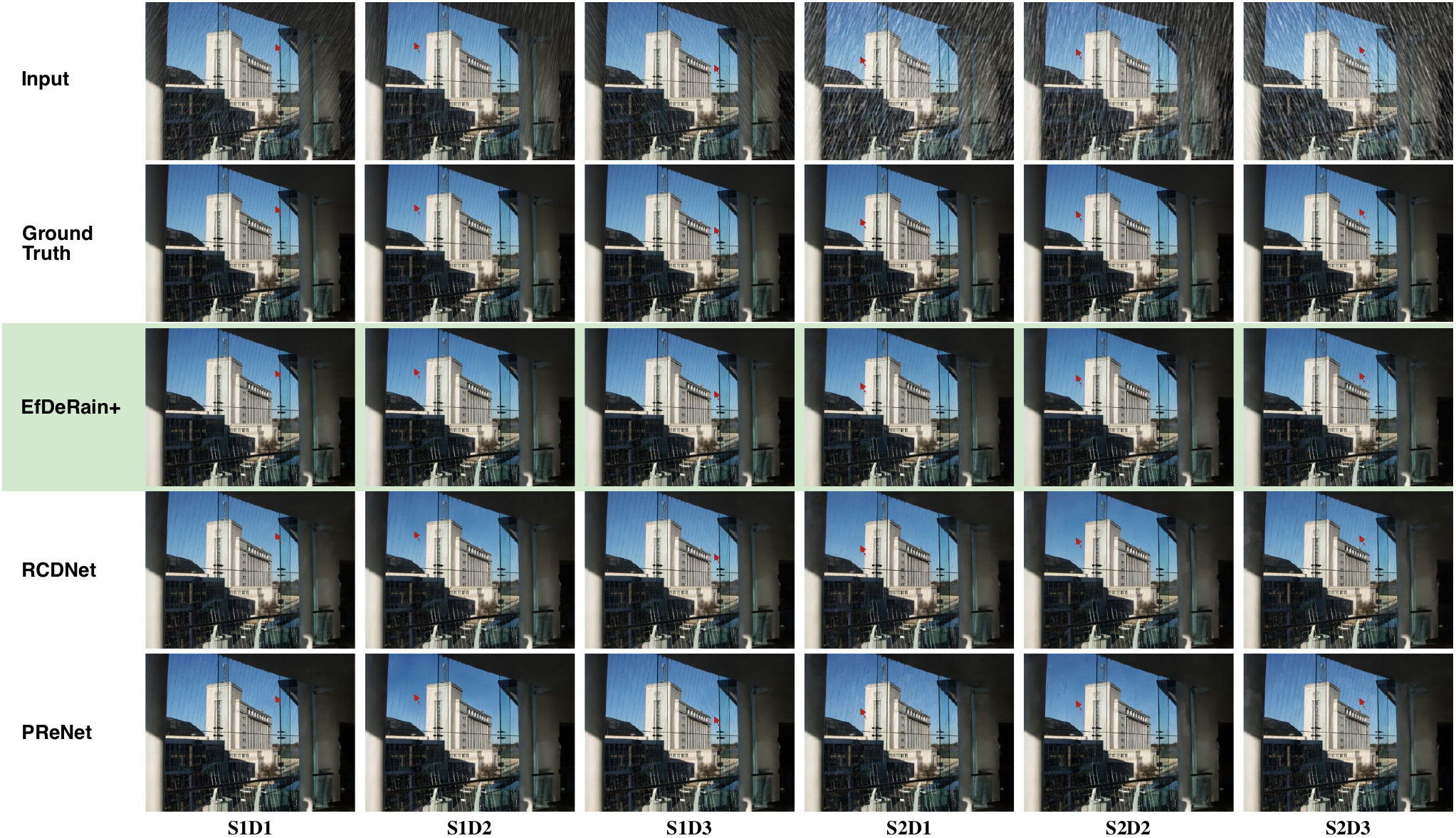}
\vspace{-20pt}
\caption{Visualization comparison with two state-of-the-art methods, \ie, RCDNet \cite{wang2020a} and PReNet \cite{ren2019cvpr}, on one case from Rain1400 with three different rain directions and two rain strengths, which are represented as S$s$D$d$ where $s$ and $d$ are the strength and direction indexes, respectively.  }
\vspace{-5pt}
\label{fig:rain1400}
\end{figure*}

{\bf Baselines.}
To be comprehensive, we perform a large-scale evaluation to compare with 10 state-of-the-art deraining methods for the derain streak task (removing rain streak), including dual attention-in-attention model (D-DAM) \cite{zhang2021dual}, rain convolutional dictionary network (RCDNet) \cite{wang2020a}, conditional variational image deraining (CVID) \cite{cvid}, 
joint rain detection and removing (JORDERE) \cite{yang2019joint},
spatial attentive deraining method (SPANet) \cite{wang2019spatial}, progressive image deraining network (PReNet) \cite{ren2019cvpr},
semi-supervised transfer learning for rain removal (SIRR) \cite{wei2019cvpr},
recurrent squeeze-and-excitation context aggregation net (RESCAN) \cite{li2018recurrent}, 
deep detail network (DDN) \cite{fu2017removing}, and Clear \cite{fu2017tip}. 
Furthermore, for deraindrop task (\ie, removing train drop) on the Raindrop dataset \cite{qian2018cvpr}, we compare 7 state-of-the-art methods \cite{li2019cvpr} as baselines, including GMM \cite{li2016rain}, JORDER \cite{yang2017cvpr}, DDN \cite{fu2017removing}, CGAN \cite{zhang2019tcsvt}, DID-MDN \cite{zhang2018iccv}, DeRaindrop \cite{qian2018cvpr}, and D-DAM \cite{zhang2021dual}.
In terms of the video deraining, we compare our method with S2VD \cite{yue2021semi}, SLDNet \cite{Yang_2020_CVPR}, SpacCNN \cite{chen2018robust}, PReNet \cite{ren2019cvpr}, DDN \cite{fu2017removing}, and FastDerain \cite{Jiang2017fastderain}.
Note that, the time cost of all compared method are one-by-one evaluated on the same server with the Intel Xeon CPU (E5-1650) and NVIDIA Quadro P6000 GPU. Some of the works, \eg, D-DAM and SLDNet, do not release the code or model and their running times are ignored.

\subsection{Comparison Results on Single-Image Deraining}
\label{subsec:comparison_image}
\subsubsection{Results on Rain100H dataset}

We report the results of 10 baseline methods and our method on Rain100H dataset in \tableref{tab:rain100h} and see that: \ding{182} Our method achieves the highest PSNR and SSIM among all compared methods. In particular, \ourmethod{} reaches 33.06 PSNR and 0.9361 SSIM while the best baseline (\ie, RCDNet \cite{wang2020a}) has 31.28 PSNR and 0.9093 SSIM. Our method achieves 5.7\% and 2.94\% relative improvements on PSNR and SSIM, respectively. \ding{183} In terms of the efficiency, our method takes an average 6.3~ms for deraining, which is about 75.6 times faster than RCDNet. The second fastest method is SPANet \cite{wang2019spatial} that runs at an average 10.4 ms and is over two times slower than our method. Moreover, SPANet gets 25.11 PSNR and 0.8332 SSIM, which are much lower than the results of \ourmethod{}.
Overall, our method can not only achieve the best image quality but also run at the fastest speed.

In terms of the qualitative evaluation, we show four challenging cases from Rain100H in \figref{fig:rain100h} and compare our method with RCDNet \cite{wang2020a} and PReNet \cite{ren2019cvpr}. Note that, we choose these two methods since their official implementations are able to generate high-quality deraining results that are consistent with their quantitative results. As shown in \figref{fig:rain100h}, we see that: \textit{First}, \ourmethod{} is able to remove the rain streaks across diverse scenes effectively. However, the RCDNet and PReNet still have some residual rain traces in the derained images (See the first case in \figref{fig:rain100h}). \textit{Second}, our method can recover the detailed structures like the fish scales and the hair in the second and the fourth cases, respectively. In contrast, the two state-of-the-art methods lead to distorted or blurred structures. For example, in the third cases, RCDNet and PReNet blur the wooden column while our method produces clear and sharp results that are consistent to the ground truth.

\subsubsection{Results on Rain1400 dataset}

\begin{table*}[t]
	\centering
    \setlength\tabcolsep{3pt}
	\caption{Experimental results on Rain1400 dataset \cite{fu2017removing}. 
	We highlight the top results with bold fonts.
	}
		\vspace{-5pt}
	\label{tab:rain1400}
	\begin{tabular}{l|llllllllll}
		\toprule
		 &  Clear \cite{fu2017tip}  & DDN \cite{fu2017removing} & RESCAN \cite{li2018recurrent} & SIRR \cite{wei2019cvpr} & PReNet \cite{ren2019cvpr} & SPANet \cite{wang2019spatial} & JORDERE \cite{yang2019joint} & CVID \cite{cvid} & RCDNet \cite{wang2020a} & \ourmethod{} \\ 
		 \midrule
		PSNR &  26.21       &   28.45    &   32.03     &    28.44   &   32.55    &  29.85  & 32.00  & 28.96  & 33.04  & \first{33.99} \\ 
		SSIM &  0.8951      &  0.8888    &   0.9314    &    0.8893  &   0.9459   &  0.9148 & 0.9347 & 0.9375 & 0.9472 & \first{0.9482} \\
		Time~(ms) &  273.7  &   57.8    &   21.4       &    393.3   &   57.4     &  10.8   & 14.3   & 246.1  & 430.3  & \first{7.0} \\
		\bottomrule
	\end{tabular}
		\vspace{-10pt}
\end{table*}

We present the comparison results on Rain1400 dataset in \tableref{tab:rain1400}. \ourmethod{} also achieves the best restoration quality with the highest PSNR and SSIM values (\ie, 33.99 and 0.9482) while running the fastest among all baselines with an average 7.0 ms. Although the state-of-the-art method RCDNet has slightly lower PSNR and SSIM than our method, it runs about 61.5 times slower than \ourmethod{}.

Rain1400 dataset synthesizes rainy images by adding rain streaks with different directions and severities. As a result, the dataset could evaluate the robustness of deraining methods to some extent. 
We take one of the scenes in Rain1400 as an example, which are added rain streaks with two different severities and three different directions. As a result, we acquire six rainy images for the same scene (See \figref{fig:rain1400}) where we denote each case as `S$s$D$d$' for the $s$-th severity and $d$-th direction.
We use \ourmethod{}, RCDNet \cite{wang2020a} and PReNet \cite{ren2019cvpr} to address all rainy images and compare their results in \figref{fig:rain1400}.
We observe that our method has similar deraining results across all severities and directions where the rain streaks are all removed and the original detailed structures (\eg, the steel wires) are properly recovered. In contrast, RCDNet and PReNet have different performance under different rain patterns. Specifically, when the rain streaks have similar direction with the steel wires (See S1D2 and S2D2 in \figref{fig:rain1400}), RCDNet and PReNet are able to remove the rain streaks effectively but eliminate the steel wires. When the rain streaks have different directions \wrt the steel wires (See S1D1, S1D3, S2D1, and S2D3), RCDNet and PReNet become less effective at rain removal since there are a lot of residual rain traces in the visualization results. Overall, the results on Rain1400 demonstrate that our method can not only remove various rain streaks effectively but also preserve the original detailed structures. 

\begin{table*}[t]
	\centering
    \setlength\tabcolsep{2.5pt}
	\caption{Experimental results on real rain SPA dataset \cite{wang2019spatial}. 
    We highlight the top results with bold fonts.
	}
		\vspace{-5pt}
	\label{tab:spa}
	\begin{tabular}{l|llllllllll}
		\toprule
		 &  Clear \cite{fu2017tip}  & DDN \cite{fu2017removing} & RESCAN \cite{li2018recurrent} & SIRR \cite{wei2019cvpr} & PReNet \cite{ren2019cvpr} & SPANet \cite{wang2019spatial} & JORDERE \cite{yang2019joint} & RCDNet \cite{wang2020a} & DualGCN \cite{fu2021rain} & \ourmethod{} \\ 
		 \midrule
		PSNR &  34.39       &   36.16    &   38.11     &    35.31   &   40.16    &  40.24  & 40.78  & 41.47  & 44.18  & \first{44.25} \\ 
		SSIM &  0.9509      &   0.9463   &   0.9707    &    0.9411  &   0.9816   &  0.9811 & 0.9811 & 0.9834 & 0.9902 & \first{0.9904} \\
		Time~(ms) &  79.4   &   32.39    &   22.67     &    371.0   &   73.6     &  10.4   & 17.6   & 434.4  &  655.1     & \first{6.9} \\
		\bottomrule
	\end{tabular}
	\vspace{-5pt}
\end{table*}

\subsubsection{Results on SPA dataset}
In terms of the experiments on real rainy images (\eg, SPA dataset \cite{wang2019spatial}), we report the results in \tableref{tab:spa}. Since the rain streaks in the rainy images of SPA is not heavy, all compared methods can achieve significantly high PSNR and SSIM values. For example, PReNet, SPANet, JORDERE, RCDNet, and DualGCN get high PSNRs with 40.16, 40.24, 40.78, 41.47, and 44.18, respectively. Our final implementation is able to achieve 44.25 and 0.9904 PSNR and SSIM where both are the highest scores among all compared methods. Moreover, \ourmethod{} runs at an average 6.9 ms per image, which is much faster than all baseline methods. 

\begin{figure}[t]
\centering
\includegraphics[width=1.0\linewidth]{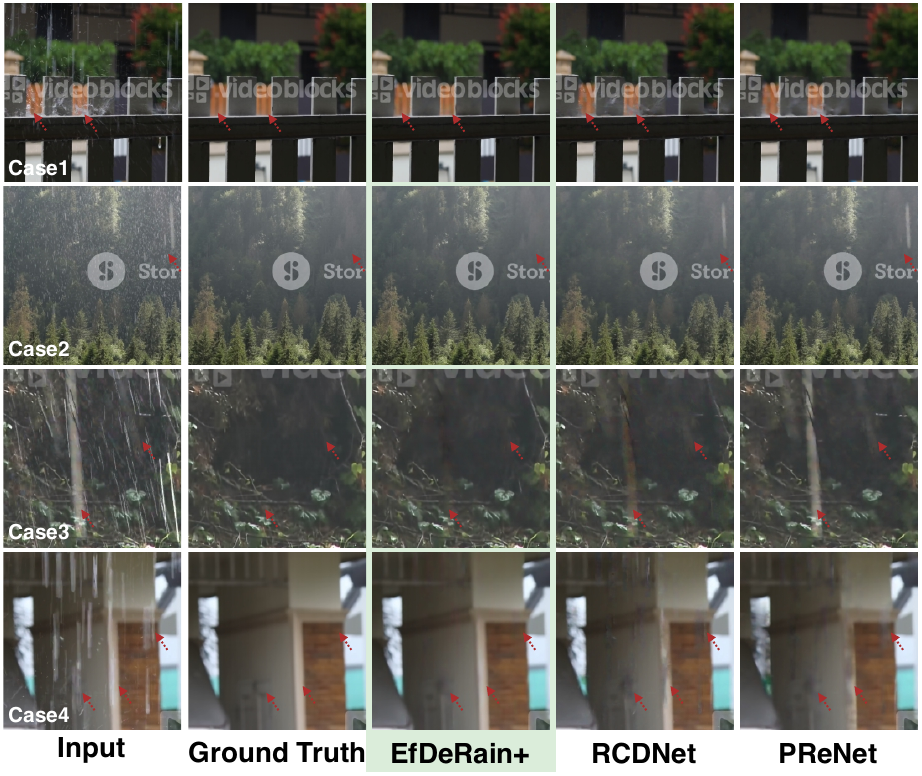}
\vspace{-20pt}
\caption{Visualization comparison with RCDNet \cite{wang2020a} and PReNet \cite{ren2019cvpr}, on four cases from SPA.  }
\vspace{-15pt}
\label{fig:spa}
\end{figure}

In addition to the quantitative evaluation, we compare four qualitative cases in \figref{fig:spa}. We see that: \ding{182} Our method is able to address different rain patterns effectively. For example, in the Case1, the splashes of rain can not be addressed by RCDNet and PReNet while \ourmethod{} is able to remove the streaks and splashes simultaneously. In the Case2, the two baseline methods can remove slight rain streaks effectively while failing to handle the heavy rain streaks. In contrast, our method with multi-scale strategy could address the two kinds of rain streaks effectively. \ding{183} RCDNet and PReNet usually leave behind some artifacts while our method can recover the detailed structures clearly (See Case3 and Case4 in \figref{fig:spa}).

\begin{table*}[t]
	\centering
    \setlength\tabcolsep{7.5pt}
	\caption{Experimental results on Raindrop dataset \cite{qian2018cvpr}. We highlight the top three results with red and green, respectively. Note that, the time cost of GMM is based on CPU implementation and is much higher than other methods.}
	\vspace{-5pt}
	\label{tab:raindrop}
	\begin{tabular}{l|llllllll}
		\toprule
		 &  GMM \cite{li2016rain}  & JORDER \cite{yang2017cvpr} & DDN \cite{fu2017removing} & CGAN \cite{zhang2019tcsvt} & DID-MDN \cite{zhang2018iccv} & DeRaindrop \cite{qian2018cvpr} & D-DAM \cite{zhang2021dual} & \ourmethod{} \\ \midrule
		PSNR &   24.58   &  27.52   &   25.23    &   21.35   &  24.76  & \first{31.57}  & 30.63  & \second{31.32} \\ 
		SSIM &   0.7808  &  0.8239  &   0.8366   &   0.7306  &  0.7930 & 0.9023 & \second{0.9268} & \first{0.9375} \\ 
		Time~(ms) &  ${1.2\times 10^5}^{*}$  &  83.4  &   54.6   &   \second{54.3}  &  150.4 & 197.8 & - & \first{7.4} \\ 
		\bottomrule
	\end{tabular}
	\vspace{-15pt}
\end{table*}

\subsubsection{Results on Raindrop dataset}

\begin{figure}[t]
\centering
\includegraphics[width=1.0\linewidth]{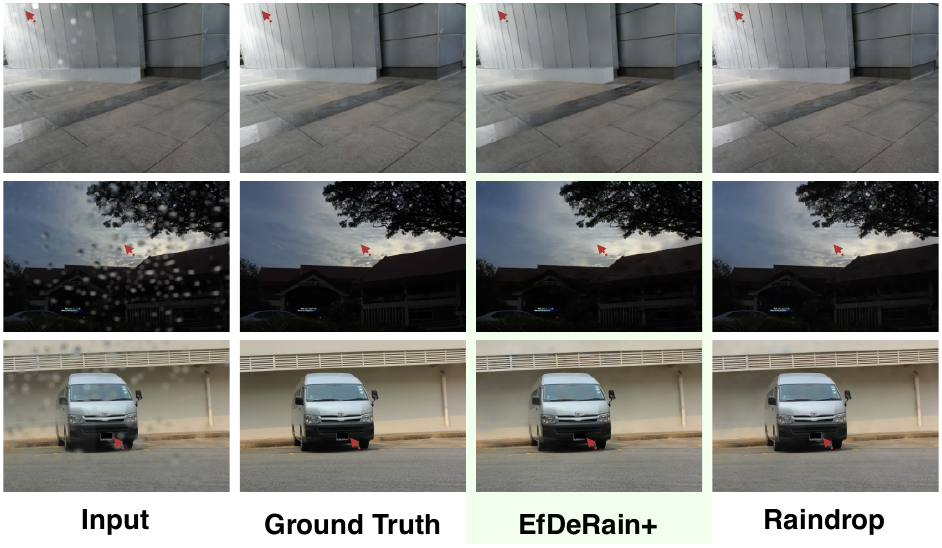}
\vspace{-20pt}
\caption{Visualization comparison with the state-of-the-art method, \ie, DeRaindrop \cite{qian2018cvpr}, on three cases from the Raindrop dataset \cite{qian2018cvpr}.}
\vspace{-10pt}
\label{fig:raindrop}
\end{figure}

Different from other datasets, Raindrop dataset corrupts images via rain drops instead of rain streaks. As shown in \figref{fig:raindrop}, the rain drops have totally different patterns with rain streaks. Among all baseline methods, DeRaindrop \cite{qian2018cvpr} and D-DAM \cite{zhang2021dual} are specifically designed for rain drop removal. According to the results in \tableref{tab:raindrop}, our method reaches the best SSIM (\ie, 0.9375) and second best PSNR (\ie, 31.32) while running the fastest with an average 7.4 ms per image. In particular, compared with the latest method D-DAM, our method presents significant advantages on both PSNR and SSIM (\ie, 31.32 vs. 30.63 and 0.9375 vs. 0.9268). These results demonstrate the effectiveness and advantages of our method on both deraining capability and efficiency.

Regarding the visualization comparison, we see that our method is able to remove rain drops effectively and recovers the original structures. In contrast, the state-of-the-art method DeRaindrop usually leads to some residual drops (See the effects indicated by red arrows in \figref{fig:raindrop}).

%
\subsection{Comparison Results on Video Deraining}
\label{subsec:comparison_video}

\begin{figure}[t]
\centering
\includegraphics[width=1.0\linewidth]{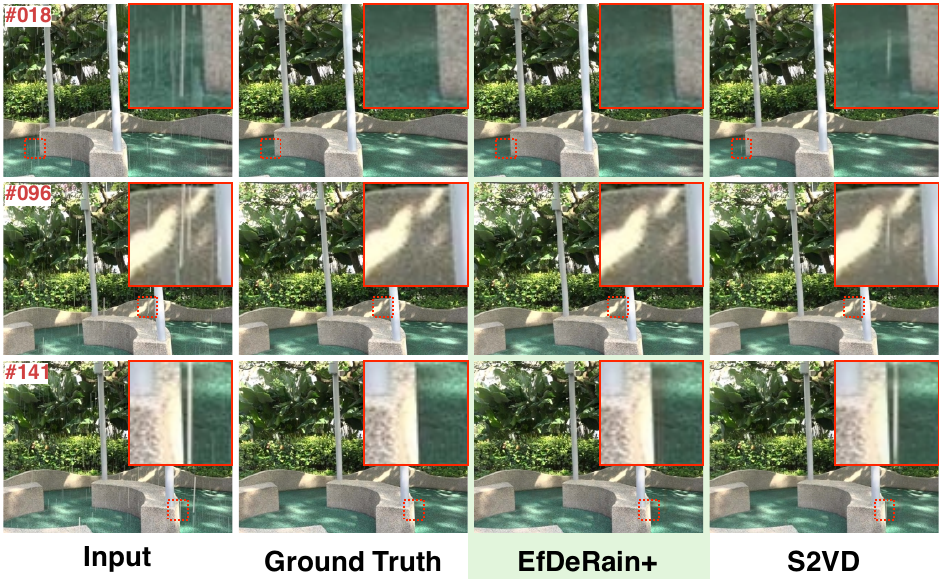}
\vspace{-20pt}
\caption{Visualization comparison with the state-of-the-art method, \ie, S2VD \cite{yue2021semi} on NTURain dataset \cite{chen2018robust}. We indicate the frame index at the top right-hand corner.}
\label{fig:nturain}
\vspace{-15pt}
\end{figure}

In addition to the single-image deraining, the proposed method is also able to remove rain in videos. We can naively use our method to derain each frame of a video and compare the results with the state-of-the-art video deraining methods. As reported in \tableref{tab:nturain}, our method achieves the highest SSIM score (\ie, 0.9713) among all compared baseline methods while having the second best PSNR with 36.98 which is slightly smaller than the result of S2VD \cite{yue2021semi}. In terms of the visualization results in \figref{fig:nturain}, we see that the best baseline method (\ie, S2VD) fails to handle some heavy rain streaks across frames while our method is able to remove diverse rain patterns.

\begin{table*}[t]
	\centering
    \setlength\tabcolsep{11pt}
	\caption{Experimental results on NTURain dataset \cite{qian2018cvpr}. The `Time' means the cost per video. 
    We highlight the top results with bold fonts.
	}
	\label{tab:nturain}
	\vspace{-5pt}
	\begin{tabular}{l|lllllll}
		\toprule
		 &  FastDerain \cite{li2016rain}  & DDN \cite{fu2017removing} & PReNet \cite{ren2019cvpr} & SPAC-CNN \cite{chen2018robust}  & SLDNet \cite{Yang_2020_CVPR} & S2VD \cite{yue2021semi} & \ourmethod{} \\ \midrule
		PSNR & 30.54   &  32.87  & 32.99 & 33.11 &  34.89  & \first{37.37}  & \second{36.98} \\ 
		SSIM &   0.9255  &  0.9497  & 0.9519 & 0.9475 &  0.9540 & \second{0.9683} & \first{0.9713} \\ 
		Time~(s) & 38.74  &  7.81  &  19.06  &  142.05  &  - & 0.05 & 0.21 \\ 
		\bottomrule
	\end{tabular}
	\vspace{-10pt}
\end{table*}

\subsection{Ablation Study}
\label{subsec:ablation_study}

We have validated our three main contributions, \ie, uncertainty-aware cascaded predictive filtering (UC-PFilt), weight-sharing multi-scale dilated filtering (WS-MS-DFilt), and RainMix, in \secref{sec:analysis}, respectively. In this section, we conduct an extensive ablation experiments to study which component contributes more to our final implementation. Specifically, we remove each contribution from our final implementation and obtain a total of eight variants (See the `Variants' column in \tableref{tab:ablationstudy}). Then, we train these models via the same training hyper-parameters as detailed in \secref{subsec:basic_impl} and test them on the Rain100H dataset. We compare different variants in \tableref{tab:ablationstudy}. When we remove UC-PFilt, WS-MS-DFilt, and RainMix from \ourmethod{}, respectively, we get three variants (See the third row to the fifth row) that contains two of our contributions and see that PSNR and SSIM decrease significantly. 
In particular, RainMix and UC-PFilt make larger contributions to our method than WS-MS-DFilt due to the higher performance drops. When we further remove a component from the above variants, we get three other variants (See the sixth row to the eighth row). Comparing these variants with the baseline method (\ie, the last row), we see that all the contributions improve the PSNR and SSIM, which demonstrates the effectiveness of the three components.
%

\begin{table}[t]
	\centering
    \setlength\tabcolsep{2.0pt}
	\caption{Ablation study on Rain100H dataset \cite{yang2017cvpr}.}
	\vspace{-5pt}
	\label{tab:ablationstudy}
	\begin{tabular}{lll|llll}
		\toprule
        \multicolumn{3}{c|}{Variants} & \multirow{2}{*}{PSNR} & \multirow{2}{*}{SSIM} & \multirow{2}{*}{Time (ms)} \\
        UC-PFilt & WS-MS-DFilt & RainMix & & & \\
        \midrule
        \ding{51} & \ding{51} & \ding{51} & 34.57 & 0.9513 & 6.26 \\
        \midrule
        \ding{51} & \ding{51} & \ding{55} & 32.09 & 0.9278 & 6.25 \\
        \ding{51} & \ding{55} & \ding{51} & 33.06 & 0.9361 & 6.25 \\
        \ding{55} & \ding{51} & \ding{51} & 32.25 & 0.9248 & 4.15 \\
        \midrule
        \ding{51} & \ding{55} & \ding{55} & 32.42 & 0.9303 & 6.22 \\
        \ding{55} & \ding{51} & \ding{55} & 31.47 & 0.9210 & 4.15 \\
        \ding{55} & \ding{55} & \ding{51} & 32.69 & 0.9271 & 4.14 \\
        \midrule
        \ding{55} & \ding{55} & \ding{55} & 31.39 & 0.9199 & 4.14 \\
		\bottomrule
	\end{tabular}
	\vspace{-10pt}
\end{table}

\section{Conclusion}

We proposed a novel deraining method denoted as \emph{\ourmethod{}}. Our method can not only achieve significantly high recovery quality but runs over 74 times more efficiently than the best state-of-the-art method. Four major contributions are beneficial to the results: \textit{First}, we formulated the deraining as a predictive filtering task and proposed the \textit{spatially-variant predictive filtering (SPFilt)} that can predict proper kernels via a deep network for the reconstruction of different pixels.
\textit{Second}, we proposed the \textit{uncertainty-aware cascaded predictive filtering (UC-PFilt)} that utilizes the predicted kernels to identify the difficulties of recovering different pixels and remove the residual rain traces effectively.
\textit{Third}, we proposed and designed an efficient multi-scale module, \ie, \textit{weight-sharing multi-scale dilated filtering (WS-MS-DFilt)}, where each pixel is filtered by multi-scale kernels dilated from a predicted single-scale kernel. As a result, the rain streaks with different sizes and strengths can be properly and automatically addressed. 
\textit{Fourth}, we proposed a simple yet effective data augmentation method for training deraining networks, \ie, \textit{RainMix}, which generates diverse rainy images by simulating the real-world situations and bridges the gap across diverse rain patterns and backgrounds.
We have performed extensive evaluations over 18 baselines to comprehensively validate our method on popular and challenging synthesis datasets, \ie, Rain100H and Rain1400, and real-world datasets, \ie, SPA and Raindrop, all of which demonstrate the advantages of our method in terms of both efficiency and deraining quality.






\ifCLASSOPTIONcaptionsoff
  \newpage
\fi

\bibliographystyle{IEEEtran}
\bibliography{ref}


%

\end{document}